%% file: main.tex
\documentclass[acmtog]{acmart}
\AtBeginDocument{%
  }

\renewcommand\footnotetextcopyrightpermission[1]{}

\def\etal {et~al\mbox{.}}
\usepackage{newpxtext}
\usepackage[table]{xcolor}
\usepackage{xspace}
\usepackage{booktabs}
\usepackage{multirow}
\usepackage{colortbl}
\usepackage{graphicx}
\usepackage{diagbox}
\usepackage{xurl}

\makeatletter
\newcommand{\ie}{\emph{i.e.}\@ifnextchar.{\!\@gobble}{}}
\newcommand{\eg}{\emph{e.g.}\@ifnextchar.{\!\@gobble}{}}
\newcommand{\etc}{etc\@ifnextchar.{}{.\@}}

\definecolor{myred}{rgb}{0.8, 0.2, 0.2}
\definecolor{purple}{rgb}{0.99,0.2,0.72}
\definecolor{myBlue2}{rgb}{.0, .0, 0.5}
\definecolor{myGreen}{rgb}{.2, .60, 0.2}

\makeatother
\begin{document}
\include{commands}
\title{Geometry-Grounded Gaussian Splatting}

\author{Baowen Zhang}
\affiliation{%
 \institution{Hong Kong University of Science and Technology}
 \city{Hong Kong}
 \country{Hong Kong}
}
\email{bzhangcm@connect.ust.hk}

\author{Chenxing Jiang}
\affiliation{%
 \institution{Hong Kong University of Science and Technology}
 \city{Hong Kong}
 \country{Hong Kong}
}
\email{cjiangan@connect.ust.hk}

\author{Heng Li}
\affiliation{%
 \institution{Hong Kong University of Science and Technology}
 \city{Hong Kong}
 \country{Hong Kong}
}
\email{eehengli@ust.hk}

\author{Shaojie Shen}
\affiliation{%
 \institution{Hong Kong University of Science and Technology}
 \city{Hong Kong}
 \country{Hong Kong}
}
\email{eeshaojie@ust.hk}

\author{Ping Tan}
\affiliation{%
 \institution{Hong Kong University of Science and Technology}
 \city{Hong Kong}
 \country{Hong Kong}
}
\email{pingtan@ust.hk}


\begin{abstract} 
\input{abstract}
\end{abstract}


\begin{teaserfigure}
\vspace{-0.3cm}
\raggedright{\large \textcolor{magenta}{\texttt{\href{https://baowenz.github.io/geometry_grounded_gaussian_splatting/}{https://baowenz.github.io/geometry\_grounded\_gaussian\_splatting}}}}\\
\vspace{0.1cm}
\centering
\setlength{\abovecaptionskip}{0.1cm}
\includegraphics[width=\textwidth]{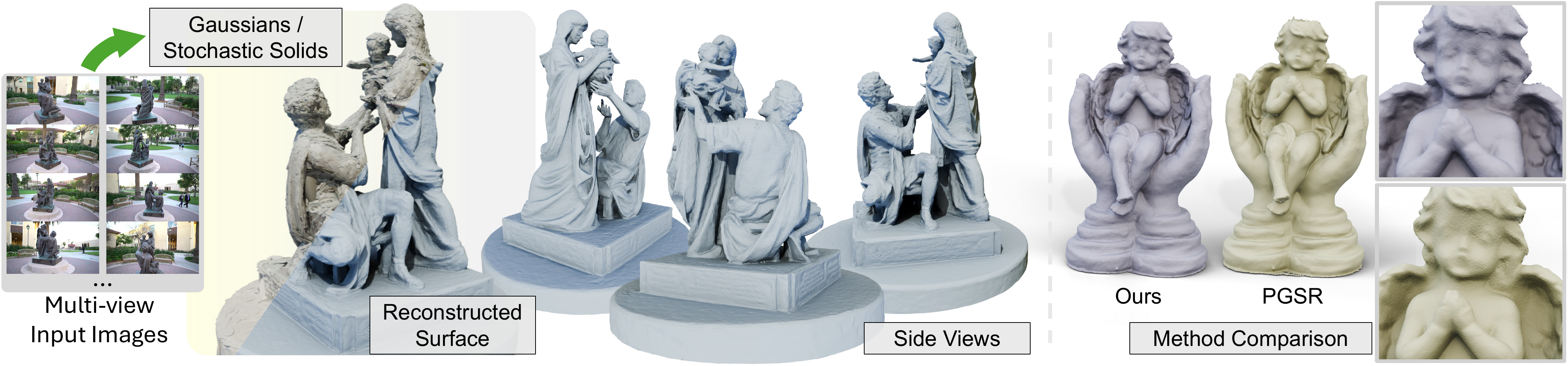}
\caption{
We prove that Gaussian primitives are equivalent to stochastic solids, and leverage this equivalence to reconstruct high-fidelity, multi-view-consistent shapes from multi-view images.
}
\label{fig:teaser}
\end{teaserfigure}

\maketitle

\section{Introduction}
\input{introduction}

\section{Related Work}
\input{related_works}

\section{Preliminary}
\input{Preliminary}

\section{Method}
\input{method}

\section{Experiments}\label{sec:exp}
\input{experiments}

\section{Conclusion}
\input{conclusion}

\bibliographystyle{ACM-Reference-Format}
\bibliography{sample-base}

\input{figure_only}

\appendix
\input{supp}

\end{document}
\endinput

%% file: commands.tex
\newcommand{\mipnerf}{Mip-NeRF 360\xspace}
\newcommand{\gs}{3D GS\xspace}
\newcommand{\ngp}{iNGP\xspace}
\newcommand{\figref}[1]{Fig.~\ref{fig:#1}\xspace}
\newcommand{\tabref}[1]{Tab.~\ref{tab:#1}\xspace}
\newcommand{\eqnref}[1]{Eqn.~\eqref{eq:#1}\xspace}
\newcommand{\appref}[1]{Appendix~\ref{app:#1}\xspace}

\newcommand{\cf}{cf.\xspace}

\newcommand{\ptd}{PTD}
\newcommand{\woptd}{w/o \ptd}

\newcommand{\FLIP}{\protect\reflectbox{F}LIP\xspace}
\newcommand{\flipview}[1]{\text{\protect\reflectbox{F}LIP}_{#1}\xspace}

\newcommand{\errmax}{\delta_{\text{max}}}
\newcommand{\erravg}{\delta_{\text{avg}}}
\newcommand{\topt}{t_{\text{opt}}}
\newcommand{\stimes}{{\times}}

\let\oldhat\hat 
\renewcommand{\vec}[1]{\mathbf{#1}} 
\renewcommand{\hat}[1]{\oldhat{\mathbf{#1}}}
\newcommand{\depth}{\zeta}

\newcommand{\ifcommentsenabled}[1]{#1}

\definecolor{mathias_color}{rgb}{.6,.4,.05}
\definecolor{michael_color}{rgb}{0,0.35,0}
\definecolor{alex_color}{rgb}{0,0,0.85}
\definecolor{markus_color}{rgb}{0,0.35,0.35}
\definecolor{bernhard_color}{rgb}{0.35,0.35,0}
\newcommand{\mathias}[1]{\ifcommentsenabled{\textcolor{mathias_color}{Mathias: #1}}}
\newcommand{\lukas}[1]{\ifcommentsenabled{\textcolor{lukas_color}{Lukas: #1}}}
\newcommand{\michael}[1]{\ifcommentsenabled{\textcolor{michael_color}{Michael: #1}}}
\newcommand{\alex}[1]{\ifcommentsenabled{\textcolor{alex_color}{Alex: #1}}}
\newcommand{\markus}[1]{\ifcommentsenabled{\textcolor{markus_color}{Markus: #1}}}
\newcommand{\bernhard}[1]{\ifcommentsenabled{\textcolor{bernhard_color}{Bernhard: #1}}}

\definecolor{edited_color}{rgb}{.7,.1,.1}
\newcommand{\new}[1]{#1} 

\definecolor{revised_color}{rgb}{.1,.1,.7}
\newcommand{\revised}[2]{#1} 

\definecolor{yellow}{rgb}{1, 1, 0.7}
\definecolor{orange}{rgb}{1, 0.85, 0.7}
\definecolor{tablered}{rgb}{1, 0.7, 0.7}
\definecolor{red}{rgb}{1, 0, 0}

\definecolor{wincolor}{rgb}{0.85, 0.0, 0.0}

\definecolor{darkyellow}{rgb}{0.8, 0.8, 0.5}
\definecolor{darkred}{rgb}{0.7, 0.3, 0.3}
\definecolor{darkgreen}{rgb}{0.3, 0.7, 0.3}
\definecolor{green}{rgb}{0, 1.0, 0}
\definecolor{pink}{rgb}{1, 0.4, 0.7}

\newcommand{\best}{\cellcolor{blue!45}}
\newcommand{\sbest}{\cellcolor{blue!30}}
\newcommand{\tbest}{\cellcolor{blue!15}}

%% file: abstract.tex
Gaussian Splatting (GS) has demonstrated impressive quality and efficiency in novel view synthesis.
However, shape extraction from Gaussian primitives remains an open problem. 
Due to inadequate geometry parameterization and approximation, existing shape reconstruction methods suffer from poor multi-view consistency and are sensitive to floaters. 
In this paper, we present a rigorous theoretical derivation that establishes Gaussian primitives as a specific type of stochastic solids. 
This theoretical framework provides a principled foundation for Geometry-Grounded Gaussian Splatting by enabling the direct treatment of Gaussian primitives as explicit geometric representations. 
Using the volumetric nature of stochastic solids, our method efficiently renders high-quality depth maps for fine-grained geometry extraction.
Experiments show that our method achieves the best shape reconstruction results among all Gaussian Splatting-based methods on public datasets.

%% file: introduction.tex
3D shape reconstruction from multi-view images is a long-standing problem with broad impact in virtual reality~\cite{snavely2006photo}, autonomous driving~\cite{r3d3}, and robotics~\cite{cadena2017past,engel2014lsd}.
Recent progress has been driven by implicit neural representations, most notably NeRF~\cite{mildenhall2020nerf}.
Many state-of-the-art methods further adopt geometry-grounded radiance fields: they start from a canonical geometry field (\eg, SDF/occupancy) and derive the rendering formulation accordingly. Methods such as VolSDF~\cite{yariv2021volume} and NeuS~\cite{wang2021neus} follow this principle by anchoring the rendering to an explicit surface and yielding reliable geometry that is consistent across views. Despite these advances, geometry-grounded radiance fields typically rely on dense sampling, \eg, ray marching, along camera rays, resulting in slow training and inference.

In contrast, Gaussian Splatting~\cite{kerbl3Dgaussians} represents scenes as a collection of Gaussian primitives and leverages efficient rasterization, enabling fast optimization and real-time novel view synthesis. Several follow-up works~\cite{huang20242d,yu2024gaussian,zhang2024rade,chen2024pgsr,guedon2025matcha,zhang2025quadratic} have extended Gaussian Splatting to shape reconstruction with promising results. 
Nevertheless, Gaussian Splatting does not inherently define a surface, unlike geometry-grounded NeRF methods that start from an SDF/occupancy field. Existing Gaussian Splatting–based methods therefore extract depth or surfaces from the Gaussian radiance field using heuristic rules. A more principled geometric formulation can improve cross-view consistency and enable higher-fidelity reconstruction, as shown in the right of Figure~\ref{fig:teaser}. 
Unlike these heuristic pipelines, we provide a principled geometric foundation for Gaussian primitives, enabling higher-fidelity shape reconstruction.

 \begin{figure}
     \centering
     \setlength{\abovecaptionskip}{0.1cm}
     \includegraphics[width=\linewidth]{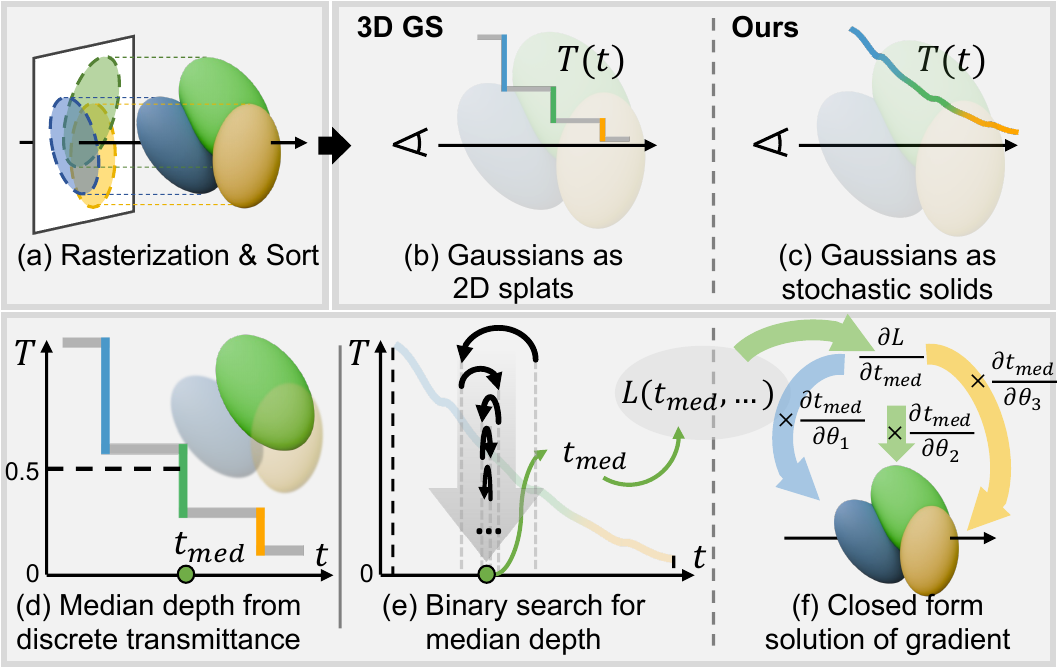}
     \caption{\textbf{Overview of our depth-rendering pipeline.} (a) We rasterize Gaussian primitives and sort them by depth. (b) Standard Gaussian Splatting yields step-wise transmittance under splat compositing. (c) Under our stochastic solid formulation, attenuation is modeled continuously within each primitive, yielding a smooth transmittance curve. (d) Prior work estimates the ray-wise median depth as the point where transmittance drops to 0.5. (e) Forward pass: we locate the median depth $t_{med}$, \emph{i.e.}, $T=0.5$, via binary search. (f) Backward pass: we backpropagate through $t_{med}$ using a closed-form gradient with respect to all Gaussians contributing to the ray.}
     \label{fig:depth_render}
 \end{figure}

In this paper, we adopt the philosophy of geometry-grounded radiance fields by equipping Gaussians with a canonical geometry field. 
We achieve this by leveraging the theoretical foundation provided by the recent work `Objects as Volumes'~\cite{Miller:VOS:2024}, which offers a stochastic interpretation of the geometry-grounded radiance field. 
Under this theory, we analyze the rendering equation of Gaussian Splatting and demonstrate that rendering a Gaussian primitive is identical to rendering a stochastic solid (Section~\ref{sec:gas}).
This unifies rendering formulations of Gaussian Splatting and NeRF-based methods, allowing us, for the first time, to derive a geometric field for Gaussian primitives. 
Using our formulation, we develop an efficient depth-rendering method that approximates the isosurface of the geometric field and extracts finer-grained geometry from Gaussian primitives (Section~\ref{sec:depth}), exhibiting inherent multi-view consistency and robustness to floaters.

Figure~\ref{fig:depth_render} illustrates our depth-rendering pipeline to exemplify our advantages in detail. Prior Gaussian Splatting–based methods define the median depth along a ray as the location where transmittance drops to 0.5, as shown in Figure~\ref{fig:depth_render}(d). 
However, because of discrete changes in transmittance, this method fails to capture the joint effect of overlapping Gaussians and leading to jagged depth steps.
In contrast, the stochastic solids model volumetric attenuation continuously and yield a smooth transmittance curve. Building on this, we endow Gaussian primitives with the same continuous behavior, enabling more detailed depth maps.
To compute the median depth, we exploit the monotonicity of transmittance and apply a binary search to locate the 0.5-transmittance crossing. We further derive a closed-form expression for the gradient of the median depth with respect to the parameters of all Gaussians along the ray for efficient backpropagation.

The main contributions of this paper are summarized as,
\begin{itemize}
    \item We analyze the rendering equation of Gaussian Splatting and demonstrate that the Gaussian primitives can be regarded as stochastic solids, which provides theoretical guidance for shape reconstruction from Gaussian Splatting (Section~\ref{sec:gas}).
    \item Based on this stochastic theory, we propose an efficient method for rendering and optimizing depth maps from Gaussian primitives, enabling accurate geometry extraction (Section~\ref{sec:depth}).
    \item Extensive experiments demonstrate that our method achieves the best reconstruction accuracy among Gaussian Splatting-based methods, while maintaining optimization efficiency of the Gaussian Splatting (Section~\ref{sec:exp}).
\end{itemize}

%% file: related_works.tex
\subsection{Continuous Radiance Fields}
NeRF~\cite{mildenhall2020nerf} models a scene as a continuous radiance field, typically parameterized by an MLP, and has shown strong performance in challenging effects such as reflections and scattering~\cite{tang2024uwnerf,levy2023seathru,ramazzina2023scatternerf}. 
Building on this backbone, Mip-NeRF introduces an anti-aliased multiscale formulation through conical-frustum rendering~\cite{barron2021mipnerf}, and Mip-NeRF~360 extends it to unbounded scenes with specialized parameterization and regularization~\cite{barron2022mipnerf360}. 
To improve efficiency, several works replace MLP ray marching with explicit volumetric parameterizations, \eg, Plenoxel's voxel-grid optimization~\cite{fridovich2022plenoxels} and SVRaster's real-time rasterization of adaptive sparse voxels~\cite{svraster}; Instant-NGP further accelerates training with multiresolution hash grids~\cite{mueller2022instant}.

While NeRF was originally designed for view synthesis, recovering accurate geometry from a generic density field is non-trivial, motivating surface-aware formulations that couple volume rendering with implicit surfaces. 
VolSDF~\cite{yariv2021volume}, NeuS~\cite{wang2021neus}, and UNISURF~\cite{oechsle2021unisurf} parameterize density through a signed distance function (SDF) and design rendering weights to obtain more faithful surfaces. 
Neuralangelo further combines multiresolution hash-grid encodings with neural surface rendering to achieve high-fidelity reconstruction from RGB captures~\cite{li2023neuralangelo}. 
GeoSVR~\cite{li2025geosvr} explores explicit sparse voxels for geometrically accurate surface reconstruction, leveraging uncertainty-aware depth constraints and voxel surface regularization to improve detail and completeness.
On the theoretical side, Objects as Volumes~\cite{Miller:VOS:2024} provides a stochastic-geometry view of representing opaque solids as volumes and clarifies when exponential transmittance-based models are physically consistent, offering principled insights into surface-oriented volume rendering. 
Although these methods can reconstruct high-quality geometry, they generally suffer from extreme time consumption.

\subsection{Primitive Based Representations}
Gaussian Splatting~\cite{kerbl3Dgaussians} represents 3D scenes using a set of 3D Gaussian primitives. Combining with rasterization techniques, it avoids the time-consuming ray marching process in NeRF rendering. As a result, it achieves both real-time rendering and accelerated training. Building on this foundation, Mip-Splatting~\cite{yu2023mip} addresses aliasing by incorporating low-pass filters, while LightGaussian~\cite{fan2024lightgaussian} optimizes memory usage with a compact representation. VastGaussian~\cite{Lin_2024_CVPR} further extends Gaussian Splatting to larger-scale scenes.
StochasticSplats~\cite{kv2025stochasticsplats} adopts a Monte Carlo estimator to enable sort-free rendering, further improving rendering efficiency.

Although 3DGS achieves high-quality novel-view synthesis, the geometry recovered from purely photometric optimization is often unreliable. To improve surface reconstruction, prior work either imposes stronger geometric priors or adds geometric supervision. SuGaR~\cite{guedon2023sugar} and NeuGS~\cite{chen2023neusg} favor surface-aligned (flattened) Gaussians to better capture object boundaries and facilitate mesh extraction. Related approaches~\cite{huang20242d,Zhang_2025_ICCV} replace 3D Gaussians with 2D primitives to encourage surface-like representations, although such constraints may reduce modeling flexibility and become unstable in complex scenes. GFSGS~\cite{jiang2025geometry} further leverages stochastic solids to construct 2D surfels for shape reconstruction. Beyond primitive design, 3DGSR~\cite{lyu20243dgsr} and GSDF~\cite{yu2024gsdf} jointly optimize Gaussians with an implicit neural SDF field, improving reconstruction fidelity while retaining splatting efficiency, and PGSR~\cite{chen2024pgsr} adds multi-view geometric regularization.

Despite promising empirical progress, geometry extraction in Gaussian Splatting still relies on heuristic depth definitions. These heuristics often yield noisy depth maps that have poor consistency across viewpoints and thus a weaker supervisory signal for optimization. This raises a fundamental question of whether Gaussian representations support an intrinsic notion of geometry akin to NeRF-based methods. We address it by adopting a stochastic approach to compute depth maps in a more principled manner for high-quality shape reconstruction.

%% file: Preliminary.tex
\subsection{Gaussian Splatting}
We first briefly revisit Gaussian Splatting (GS). A 3D Gaussian primitive is defined as follows:
\begin{equation}
    G(\mathbf{x}) = oe^{-(\mathbf{x}-\mathbf{x}_c)^\top \mathbf{\Sigma}^{-1}(\mathbf{x}-\mathbf{x}_c)},
\end{equation}
where $o$ is the opacity, $\Sigma\in\mathbf{R}^{3\times 3}$ is the covariance, $\mathbf{x}\in\mathbf{R}^3$ represents a point in 3D space, and $\mathbf{x}_c\in\mathbf{R}^3$ denotes the Gaussian's center. To enable fast rasterization, Gaussian Splatting (GS) methods employ a local affine approximation to project 3D Gaussian primitives to 2D Gaussians on the image plane with the covariance matrix $\mathbf{\Sigma}'_{2D}$
The opacity of the 2D Gaussian $\alpha (\mathbf{u})$ is defined as the maximum value of the projected 2D Gaussian:
\begin{equation}
    \alpha(\mathbf{u}) = o e^{-(\mathbf{u}-\mathbf{u}_c)^\top \mathbf{\Sigma}_{2D}^{'-1}(\mathbf{u}-\mathbf{u}_c)},
    \label{eq:alpha}
\end{equation}
where $\mathbf{u}$ is the coordinate of the pixels in the image space, $\mathbf{u}_c$ is the projected center of the Gaussian.
In this way, 3D Gaussian primitives are projected into 2D Gaussians. These 2D Gaussians are then sorted and alpha-blended to compute the final color. More details can be found in the supplementary material.

\begin{figure*}
    \centering
    \includegraphics[width=0.95\linewidth]{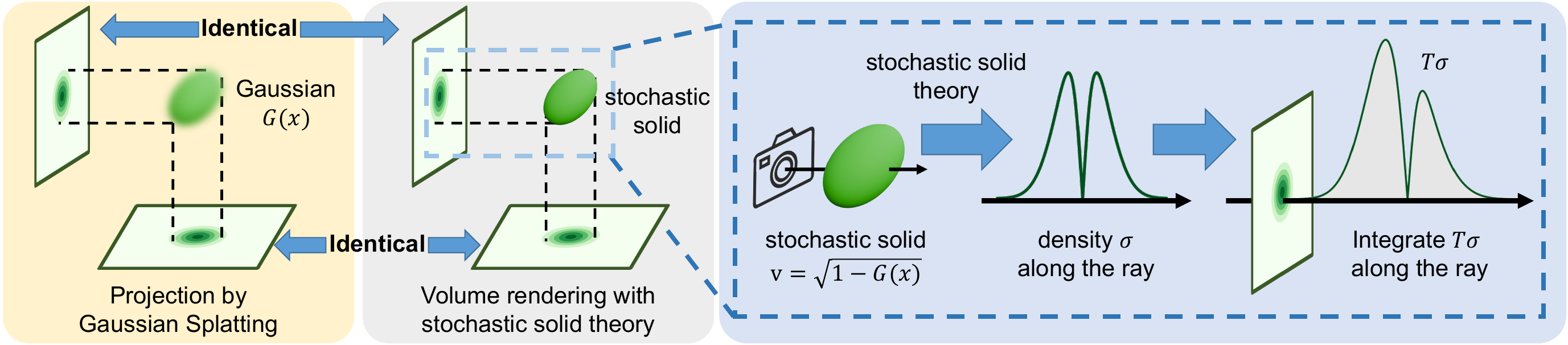}
    \caption{Given a Gaussian primitive, we regard it as a stochastic solid and derive an appropriate attenuation coefficient with Equation~\ref{eq:oav}. With our attenuation coefficient, the volume rendering of this stochastic solid is equivalent to the rasterization rendering developed in the original Gaussian Splatting.}
    \label{fig:single_gaussian}
\end{figure*}

 \subsection{Objects as Volumes}
 \label{sec:oav}
In this subsection, we provide a brief overview of~\cite{Miller:VOS:2024}, which presents a method to render stochastic solids using volume rendering. For a stochastic opaque solid characterized by its occupancy $\mathcal{O}$ and vacancy $\mathrm{v}$, \ie, $1-\mathcal{O}$, the authors derive the attenuation coefficient $\sigma$ of the object as follows:
 \begin{equation}
     \sigma(\mathbf{x}, \mathbf{\omega}) = \lvert \mathbf{\omega} \cdot \nabla log(\mathrm{v}(\mathbf{x})) \rvert=\frac{\lvert \mathbf{\omega} \cdot \nabla \mathrm{v}(\mathbf{x}) \rvert}{\mathrm{v}(\mathbf{x})},
     \label{eq:oav}
 \end{equation}
where $\mathbf{\omega}$ is the viewing direction and $\mathbf{x}$ is the 3D position.
With this attenuation coefficient, they derive the volume rendering for a stochastic solid as,
 \begin{equation}
 \begin{gathered}
     \mathbf{C} = \int_{t_{n}}^{t_{f}} p(t)\mathbf{c}(\mathbf{x}(t),\mathbf{\omega}) \, dt,\\
     p(t) = T(t)\sigma(\mathbf{x}(t),\mathbf{\omega}),\\
    T(t)  = exp\left(-\int_{t_{n}}^{t} \sigma(\mathbf{x}(s),\mathbf{\omega}) \, ds\right),\\
 \end{gathered}
     \label{eq:volume}
 \end{equation}
 where $p$ is the free-flight distribution~\cite{Miller:VOS:2024} that represents the statistical distribution of the distances that the light travels before collision and serves as the weight for color integration, and $T(t)$ is the transmittance along the ray. 

 In our work, we regard a 3D Gaussian primitive as a stochastic solid and design an appropriate attenuation coefficient $\sigma$ for it.  With this coefficient, the volume rendering of a Gaussian primitive, as described in Equation~\ref{eq:volume}, is equivalent to its rasterized rendering. This enables us to study Gaussian Splatting in a more principled manner and develop a shape reconstruction method for Gaussian primitives.

%% file: method.tex
In the following sections, we first introduce our method for a single Gaussian primitive. We then design an efficient method for rendering depth maps from multiple Gaussian primitives.

\subsection{Gaussian Primitives as Stochastic Solids}\label{sec:gas}
We treat a Gaussian primitive as a stochastic solid~\cite{Miller:VOS:2024} and derive its rendering function. 
As shown in Figure~\ref{fig:single_gaussian}, we prove that, with a proper attenuation coefficient $\sigma$, the volume rendering of this stochastic Gaussian solid is equivalent to the rasterization rendering of the original Gaussian Splatting. 
Specifically, the opacity $\alpha$ for a pixel in Equation~\ref{eq:alpha} corresponds to the maximum value of the Gaussian function along that pixel's view ray (as proved in the supplementary). Therefore, the rendered color of a single Gaussian is given by:
 \begin{equation}
     \mathbf{C} = \mathbf{c}\alpha =\mathbf{c}G(t^*),
     \label{eq:single_splat}
 \end{equation}
 where $t^*$ is the maximum point along the ray $l:\mathbf{o}+\mathbf{\omega}t$, 
 and we denote $G(\mathbf{o}+\mathbf{\omega}t^*)$ by $G(t^*)$ for simplification. 

Equation~\ref{eq:single_splat} cannot uniquely determine the attenuation coefficient. So, we impose three additional constraints. Given a Gaussian primitive $G(\mathbf{x})$, we assume that
\begin{itemize}
    \item[--] When $G(\mathbf{x_1})\geq G(\mathbf{x_2})$, it follows that  $\mathrm{o}(\mathbf{x}_1)\geq\mathrm{o}(\mathbf{x}_2)$, indicating that positions closer to the Gaussian center have higher occupancy;
     \item[--] The occupancy of the solid approaches 0 when $\mathbf{x}$ is far from the Gaussian center;
    \item[--] The occupancy $\mathrm{o}(\mathbf{x})$ is differentiable from $\mathbf{x}$.
\end{itemize} This leads us to derive a straightforward and unique expression for the vacancy:
\begin{equation}
    \mathrm{v}(\mathbf{x}) = \sqrt{1-G(\mathbf{x})}.
    \label{eq:G2v}
\end{equation}

To prove Equation~\ref{eq:G2v}, we first derive the volume rendering of a Gaussian primitive following the method in ~\cite{Miller:VOS:2024} as,
\begin{equation}
    \mathbf{C} = \int_{-\infty}^{\infty} T(t)\sigma(\mathbf{x}(t),\mathbf{\omega})\mathbf{c} \, dt = \mathbf{c}(1 - \mathrm{v}(t^*)^2),
    \label{eq:v2v}
\end{equation}
where the attenuation coefficient $\sigma$ originates from the stochastic solid as described in Equation~\ref{eq:oav}, resulting in the integral in terms of vacancy $\mathrm{v}(t^*)$. Compared with the rasterization in Equation~\ref{eq:single_splat} and the volumetric rendering in Equation~\ref{eq:v2v}, we can obtain, 
\begin{equation}
    \mathbf{C} = \mathbf{c}G(t^*) =\mathbf{c}(1 - \mathrm{v}(t^*)^2),
    \label{eq:GS_eq_vol}
\end{equation}

\noindent In other words, we derive the following condition,
\begin{equation}
    \mathrm{v}(t^*) = \sqrt{1 - G(t^*)}.
\end{equation}
Therefore, a stochastic Gaussian solid can produce the same rendering results as the rasterization in Gaussian Splatting if its vacancy adheres to Equation~\ref{eq:G2v}.  
The proof of uniqueness and other details can be found in our supplementary materials. 
Now, we can use Equation~\ref{eq:oav} and Equation~\ref{eq:G2v} to obtain attenuation coefficients $\sigma$ inside a Gaussian primitive, allowing us to obtain accurate depth maps and smooth optimization.

This property lets us move beyond heuristic geometry readouts,
leading to a principled shape reconstruction approach built directly on Gaussian primitives.
In the following sections, we apply this theory to Gaussian Splatting and demonstrate that it substantially improves shape reconstruction.

 \subsection{Depth from Stochastic Solids}\label{sec:depth}
\begin{figure*}[t]
    \centering
    \includegraphics[width=\linewidth]{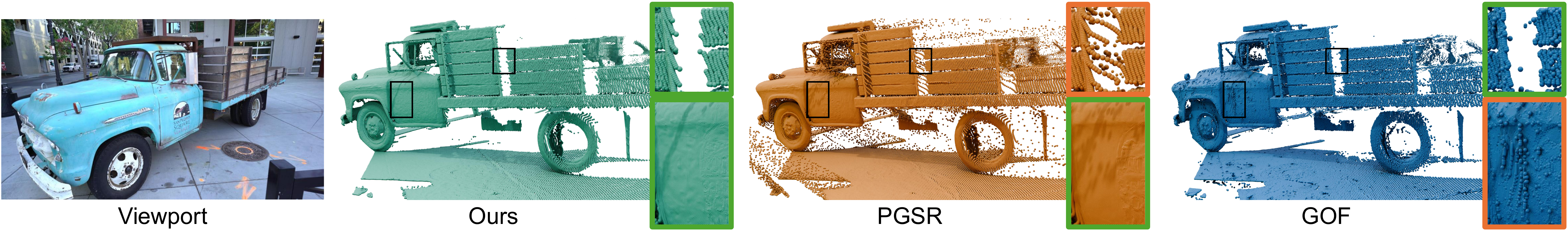}
    \caption{\textbf{Depth maps of Gaussian Splatting-based methods.} We visualize the depth maps by converting them to 3D points. Our method produces a clean and smooth depth map. PGSR uses expected depth, yielding much noise at edges. GOF uses median depth and suffers from unsmooth depth changes.}
    \label{fig:depth}
\end{figure*}

In Gaussian Splatting, photometric supervision alone is insufficient to reconstruct high-quality shapes. To better recover surface geometry, recent works~\cite{guedon2023sugar,huang20242d,chen2024pgsr} render depth maps from Gaussian primitives and add geometric regularizers, and then backpropagate their gradients to the Gaussian parameters.
Nevertheless, the rendered depth maps are noisy and have poor cross-view consistency, \eg, as shown in Figure~\ref{fig:multi_view_consistency} and \ref{fig:depth}, providing weak geometric supervision. 
This motivates us to improve depth rendering in Gaussian Splatting by utilizing attenuation coefficients derived from stochastic solids. 

The rendering pipeline is shown in Figure~\ref{fig:depth_render}.
We first derive our depth computation method, then show that it improves multi-view consistency and produces cleaner depth maps.

\begin{figure}
    \centering
    \includegraphics[width=\linewidth]{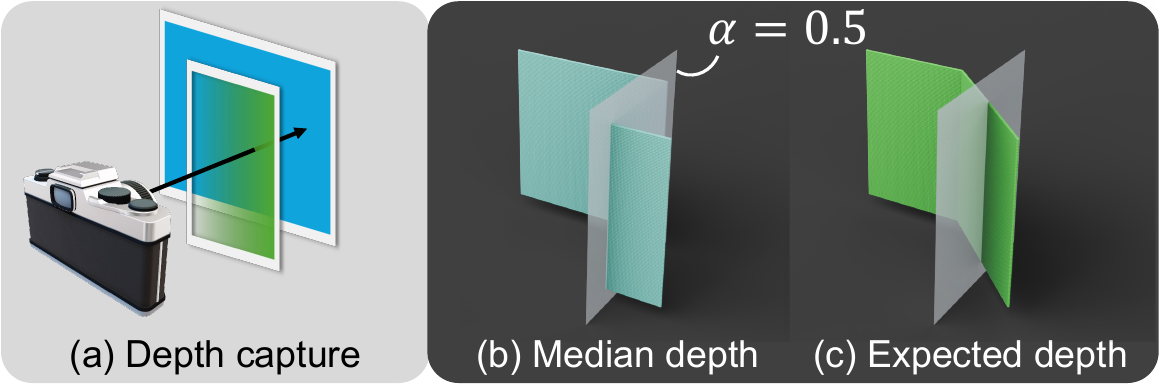}
    \caption{\textbf{Illustration of depth rendering methods.} (a) The green plane’s opacity $\alpha$ decreases smoothly from 1 on the right to 0 on the left. Consequently, (b) the median depth changes in a step-like manner, whereas (c) the expected depth varies continuously.}
    \label{fig:depth_illustration}
\end{figure}

\subsubsection{Depth definition}
Following prior Gaussian Splatting methods, we use the median depth $t_{med}$ for geometric regularization:
\begin{equation}
    t_{med} = T^{-1}(0.5),
\label{eq:t_depth}
\end{equation}
where $T^{-1}(*)$ is the inverse function of the transmittance $T(t)$.
Following prior work~\cite{Condor2024Gaussians,blanc2025raygauss,blanc2025raygaussx}, we assume that the events of a view ray intersecting different Gaussians are statistically independent. Under this assumption, the overall transmittance at $t$ along the ray is the product of the transmittance calculated at each Gaussian primitive as,
\begin{equation}
    T(t) = \prod_i T_i(t)\, ,
\label{eq:T}
\end{equation}
where $T_i(t)$ is the transmittance of the $i$-th Gaussian as:
\begin{equation}
    T_i(t) = \begin{cases}
        \mathrm{v}_i(t),\quad &t\leq t^*_i \\
        \mathrm{v}_i(t^*_i)^2 / \mathrm{v}_i(t),\quad &t>t^*_i.
        \end{cases}
    \label{eq:Ti}
\end{equation}
Here, $t^*_i$ is the Gaussian's maximum point along the camera ray.
Equation~\ref{eq:Ti} is derived from the continuous attenuation profile within each Gaussian as defined in Equation~\ref{eq:oav}, capturing more detailed geometry information.
The derivation of Equation~\ref{eq:Ti} can be found in the supplementary material.

\begin{figure}
    \centering
    \includegraphics[width=0.9\linewidth]{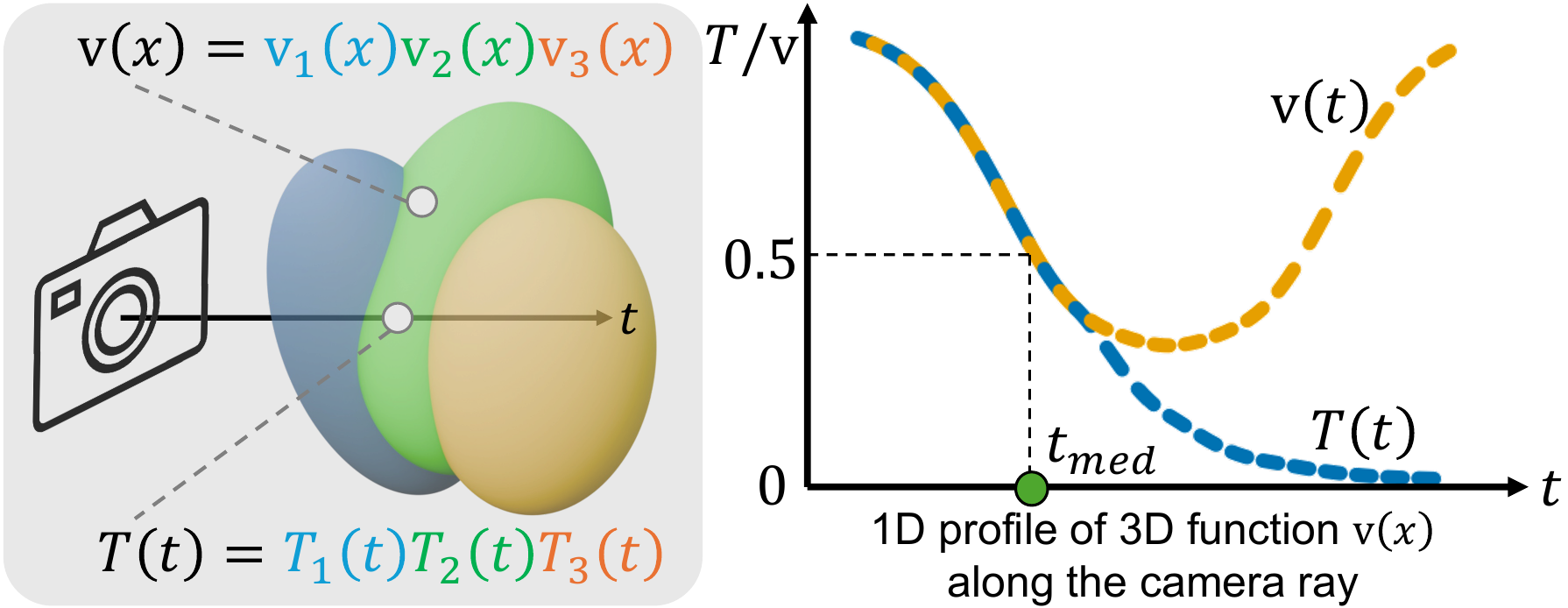}
    \caption{\textbf{Illustration of vacancy along the camera ray.} The vacancy value on the ray equals the transmittance on the front side of the Gaussians.}
    \label{fig:isosurface}
\end{figure}

\paragraph{Discussion} 
Previous methods estimate depth either from per-view depth planes~\cite{zhang2024rade,yu2024gaussian} that are view-dependent by design, or via opacity-weighted ray averaging~\cite{chen2024pgsr} that is easily biased by view-specific floaters.
These depth extraction strategies often lead to poor cross-view consistency. 
In contrast, we will show that interpreting Gaussian Splatting as a stochastic solid yields a median depth estimate with strong multi-view consistency.
Recall that the median depth is the point where the transmittance first reaches a fixed threshold, \ie, $T=0.5$. 
From Equations~\ref{eq:Ti} and~\ref{eq:T}, if the overall transmittance crossing $T=0.5$ occurs before the peak of the contributing Gaussians, then the transmittance coincides with the 3D vacancy field as illustrated in Figure~\ref{fig:isosurface}. So, the depth map is a view-independent $0.5$-level isosurface. 
This regime is common because optimization clusters high-opacity Gaussians near the surface, making the transmittance drop mainly on their near sides. While floaters can still perturb the ray, the median depth is more robust to such outliers than the alpha-averaged expected depth, leading to stronger multi-view consistency.

Beyond improved multi-view consistency, our method produces cleaner depth maps as shown in Figure~\ref{fig:depth}.
Depth obtained via alpha-weighted compositing tends to interpolate between foreground and background at boundaries, leading to blurred silhouettes as shown in Figure~\ref{fig:depth_illustration}.
Median depth, defined by the $T=0.5$ crossing, gives sharper boundary transitions.
However, in prior Gaussian Splatting formulations, transmittance is updated in discrete steps, so the 0.5 crossing often snaps to a single Gaussian; neighboring pixels may therefore select different Gaussians, producing jagged artifacts.
Our stochastic-solid formulation models attenuation continuously within each Gaussian, yielding a smooth transmittance function and reducing staircasing while preserving sharp boundaries.

\subsubsection{Implementation}
In general, Equation~\ref{eq:t_depth} does not admit a closed-form solution. To address this, we exploit the monotonicity of transmittance along each ray and use an iterative binary search to find the median depth.
During backpropagation, we do not require an iterative search. Instead, we derive the closed-form solution for the gradient of depth $t_{med}$ with respect to the Gaussians' parameters as,
\begin{equation}
    \frac{\partial t_{med}}{\partial \theta} = -\frac{\partial T(t_{med};\theta)}{\partial \theta}/\frac{\partial T(t;\theta)}{\partial t}\big|_{t=t_{med}},
\label{eq:depth_gradient}
\end{equation}
where $\theta$ denotes the Gaussian parameters along the ray.

Equation~\ref{eq:depth_gradient} shows that the gradient can be distributed to all contributing Gaussians along the ray, unlike previous methods where the gradient of the median depth was only applied to a single Gaussian.  This stems from our stochastic-solid formulation, which yields a differentiable transmittance function. As a result, the median depth $t_{med}$ varies smoothly with the Gaussian parameters, providing denser supervision for optimization.
The derivation of Equation~\ref{eq:depth_gradient} and implementation details are provided in the supplementary material.

\subsection{Optimization with Stochastic Solids}
We optimize scenes using photometric loss~\cite{kerbl3Dgaussians}, normal consistency loss~\cite{huang20242d}, and multi-view regularization~\cite{chen2024pgsr}; details are provided in the supplementary material.
These losses require rendering RGB images, normal maps, and depth maps. 
Fully volumetric rendering for all modalities is computationally expensive~\cite{Condor2024Gaussians,blanc2025raygauss,blanc2025raygaussx}. We therefore retain the standard Gaussian Splatting approximation for RGB and normals~\cite{zhang2024rade}, while computing depth using Eq.~\ref{eq:t_depth}.
Experiments show that this setting can significantly improve the shape reconstruction accuracy of Gaussian Splatting, while maintaining the efficiency.
Nevertheless, we believe that extending our volumetric formulation to RGB and normal rendering can further improve accuracy, which we leave for future work.

%% file: experiments.tex
\input{table/dtu}
\input{table/tnt}
We evaluate our method on several public datasets and compare it with existing state-of-the-art methods.

\paragraph{Implementation Details}
We use a local affine approximation and adopt RaDe-GS~\cite{zhang2024rade} to estimate each Gaussian peak $t_i^*$. For efficiency, we follow gsplat~\cite{ye2025gsplat} and use warp-level reductions for gradient accumulation. We apply the 3D filtering from Mip-Splatting~\cite{yu2023mip} (without its 2D filter), the densification strategy from GOF~\cite{yu2024gaussian}, and the exposure compensation from PGSR~\cite{chen2024pgsr}. Multi-view regularization is implemented with a custom CUDA kernel. We will release our code.

\paragraph{Datasets}
We evaluate reconstruction accuracy on DTU~\cite{jensen2014large} and Tanks \& Temples (TnT)~\cite{Knapitsch2017}. Following prior work, we use the standard 15-scene DTU split and the common 6-scene TnT subset. We report Chamfer Distance on DTU and F1-score on TnT.

\paragraph{Mesh Extraction}
Following previous works~\cite{zhang2024rade,yu2024gaussian}, we apply the TSDF fusion~\cite{curless1996volumetric} implemented by Open3D~\cite{Zhou2018} to extract meshes for the DTU dataset and adopt Marching Tetrahedra~\cite{yu2024gaussian,guedon2025milo} for large-scale scenes in the Tanks \& Temples dataset. 
Inspired by GOF, we define an indicator function over 3D space for Marching Tetrahedra. Specifically, a point is classified as inside the mesh if it is occluded in any training view, \ie, if its transmittance falls below $0.5$; otherwise, it is classified as outside.

\subsection{Reconstruction Comparison}
\begin{figure*}[t]
    \centering
    \setlength{\abovecaptionskip}{0.1cm}
    \includegraphics[width=\linewidth]{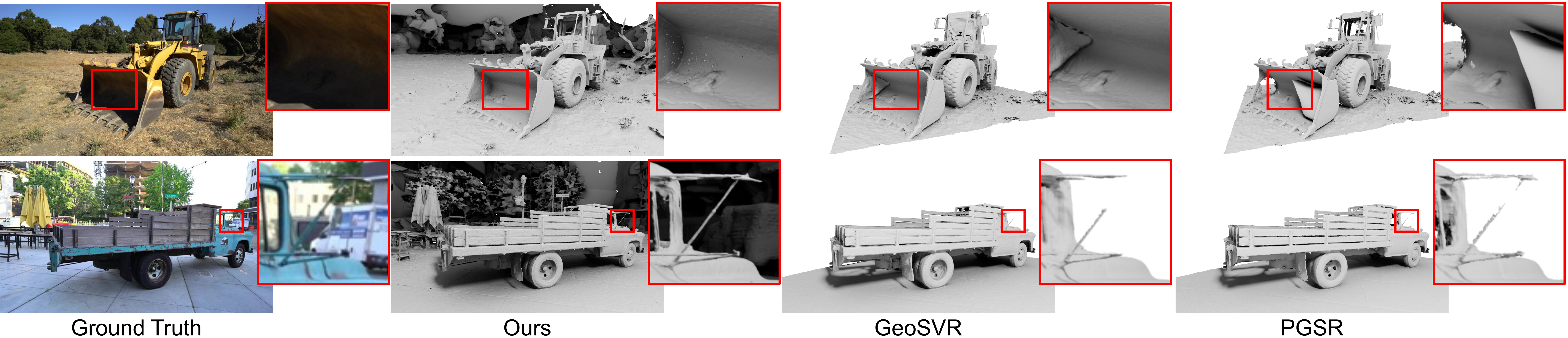}
    \caption{\textbf{Qualitative comparison on Tanks \& Temples~\cite{Knapitsch2017} dataset.} We compare our method with GeoSVR and PGSR. Our method reconstructs plausible meshes with finer geometric details}
    \label{fig:tnt_comparison}
\end{figure*}

We compare our method against existing state-of-the-art methods in the shape reconstruction task. Table~\ref{tab:dtu_result} and Table~\ref{tab:tnt} show the accuracy on the DTU and TnT datasets. 
The multi-view regularizer adopted in PGSR and GeoSVR substantially boosts DTU accuracy; using this regularization, our method achieves comparable performance to both.
In TnT, our method significantly outperforms existing Gaussian Splatting-based methods because of our depth-rendering formulation, which enables finer geometric details, enforces view-consistent geometry, and is robust to floaters. 
Figure~\ref{fig:tnt_comparison} provides qualitative comparisons among shape reconstruction methods. Our method reconstructs finer details with more accurate geometry. Additional qualitative results are shown in Figure~\ref{fig:dtu} and Figure~\ref{fig:tnt}.

We report runtimes in Table~\ref{tab:dtu_result} and Table~\ref{tab:tnt}. For the same number of iterations, our method is faster than GeoSVR (15 vs.\ 53 min.) and PGSR (25 vs.\ 30 min.), thanks to a more efficient implementation of multi-view regularization. Our runtime is higher than the fastest baselines due to the added cost of the binary search and the multi-view term. We expect further speedups by tightening the initial depth interval of the binary search, which we leave for future work.

\begin{figure*}[t]
    \centering
    \setlength{\abovecaptionskip}{0.0cm}
    \includegraphics[width=1\linewidth]{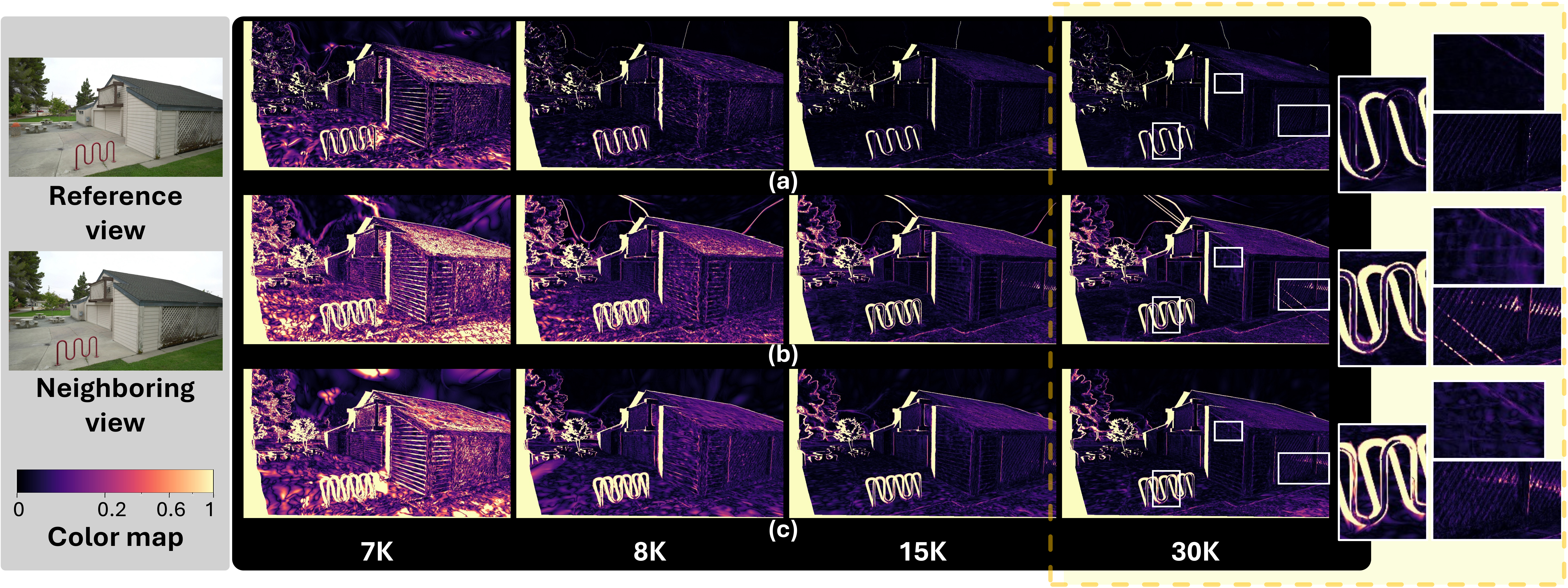}
    \caption{\textbf{Cycle reprojection error per iteration.} We visualize the cycle reprojection error between a reference view and its nearest neighboring view throughout optimization. Our method (a) attains lower errors on foreground regions and achieves full coverage faster than the other methods. PGSR (b) relies on planar-based depth accumulation, which leads to weaker multi-view consistency than our volumetric formulation and suffers from noticeable floaters. We add the multi-view regularization to RaDe-GS (c) and train it using expected depth. Zoomed-in patches are shown on the right.}
    \label{fig:multi_view_consistency}
\end{figure*}

\subsection{Multi-view Consistency}
Geometric consistency across views is essential for accurate shape reconstruction. To evaluate each depth-rendering method, we compute per-pixel cycle reprojection error during training. For a reference and neighboring view, we render depth maps of the reference view, back-project pixels to 3D, project into the neighbor to sample the corresponding depth, then back-project and reproject to the reference. The cycle error is the Euclidean distance between the original and reprojected pixel locations.

We compare our method with PGSR~\cite{chen2024pgsr} and RaDe-GS~\cite{zhang2024rade}. PGSR defines the ray-surface intersection using a plane orthogonal to the shortest axis of each 3D Gaussian, whereas RaDe-GS uses the ray-wise maximizer of the Gaussian response. For a fair comparison, we evaluate RaDe-GS augmented with the same multi-view regularization used in PGSR, and enable geometric regularization at 7K iterations for all methods. As shown in Figure~\ref{fig:multi_view_consistency}, our method yields a better initialization and converges faster, achieving the lowest reprojection error at 30K iterations, mainly due to our depth formulation based on stochastic theory. In contrast, the other methods exhibit noticeably larger reprojection errors, and floaters that arise during training further exacerbate the inconsistency.
More results can be found in Figure~\ref{fig:reprojection_more}.

\subsection{Ablation Study}
\input{table/ablation}
In this section, we evaluate the contribution of each component when integrated into our method. Table~\ref{tab:ablation} reports quantitative results on the TnT dataset. The geometric multi-view term $L_{gc}$ penalizes cycle reprojection error; however, it brings only marginal gains in our setting, because our depth-rendering formulation already provides strong multi-view consistency. In contrast, the normal consistency loss and the exposure compensation module consistently improve reconstruction quality. Finally, compared with the other two depth-rendering baselines equipped with similar regularizers, our method achieves a more accurate shape reconstruction.

%% file: table/dtu.tex
\begin{table*}[ht]
\centering
\caption{Quantitative comparison on the DTU dataset~\cite{jensen2014large}. We report Chamfer Distance and average optimization time for different methods. Among explicit Gaussian Splatting–based approaches, our method achieves the best results and attains accuracy comparable to GeoSVR. All Gaussian Splatting methods are evaluated using half-resolution images.}
\vspace{-0.4cm}
\resizebox{.98\textwidth}{!}{
\begin{tabular}{@{}llcccccccccccccccclcc}
\hline
 \multicolumn{3}{c}{} & 24 & 37 & 40 & 55 & 63 & 65 & 69 & 83 & 97 & 105 & 106 & 110 & 114 & 118 & 122 & & Mean & Time \\ \cline{4-18} \cline{20-21}
\multirow{4}{*}{\rotatebox[origin=c]{90}{implicit}} & NeRF~\cite{mildenhall2020nerf} & & 1.90 & 1.60 & 1.85 & 0.58 & 2.28 & 1.27 & 1.47 & 1.67 & 2.05 & 1.07 & 0.88 & 2.53 & 1.06 & 1.15 & 0.96 & & 1.49 & $>\text{12h}$\\
 & VolSDF~\cite{yariv2021volume} & &  1.14 &  1.26 &  0.81 & 0.49 & 1.25 & 0.70 &  0.72 &  1.29 & 1.18 &  0.70 & 0.66 & 1.08 &  0.42 &  0.61 &  0.55 & & 0.86 & $>\text{12h}$\\
 & NeuS~\cite{wang2021neus} & &  1.00 & 1.37 & 0.93 &  0.43 & 1.10 &  0.65 &  0.57 &  1.48 & 1.09 &  0.83 &  0.52 &  1.20 & 0.35 &  0.49 &  0.54 & &  0.84 & $>\text{12h}$\\
 & Neuralangelo~\cite{li2023neuralangelo} & & \tbest 0.37 & 0.72 & 0.35 & 0.35 &  0.87 & 0.54 &  0.53 &  1.29 &  0.97 &  0.73 & 0.47 & 0.74 &  0.32 & 0.41 & 0.43 & & 0.61 & $>\text{12h}$\\ 
 \cline{2-2} \cline{4-18} \cline{20-21}
\multirow{9}{*}{\rotatebox[origin=c]{90}{explicit}} 
&  3D GS~\cite{kerbl3Dgaussians} & & 2.14 & 1.53 & 2.08 & 1.68 & 3.49 & 2.21 & 1.43 & 2.07 & 2.22 & 1.75 &  1.79 & 2.55 & 1.53 & 1.52 & 1.50 & & 1.96 & \best 7.8m\\
 & 2D GS~\cite{huang20242d} &&  0.48 & 0.91 &  0.39 &  0.39 &  1.01 &  0.83 &  0.81 &  1.36 &  1.27 &  0.76  &  0.70 &  1.40 &   0.40 &   0.76 &  0.52 &&   0.80 & \tbest 11.3m \\
 & GOF~\cite{yu2024gaussian} & &  0.50 & 0.82 & 0.37 & 0.37 & 1.12 &  0.74 & 0.73 & 1.18 & 1.29 & 0.68 & 0.77 &  0.90 & 0.42 & 0.66 & 0.49 &&  0.74 & 52m\\
 & 3DGSR~\cite{lyu20243dgsr} & & 0.44 & 0.96 & 0.40 & 0.36 & 1.02 & 0.80 & 0.64 & 1.20 & 1.08 & 0.97 & 0.54 & 0.72 & 0.37 & 0.52 & 0.42 & & 0.70 & \diagbox{}{}\\
   & RaDe-GS~\cite{zhang2024rade} & & 0.43 &  0.75 & 0.35 &  0.37 & 0.81 &  0.74 & 0.74 &  1.19 & 1.20 & 0.65 & 0.61 & 0.84 & 0.35 & 0.66 &  0.46 && 0.68  & \sbest 8.2m\\
  & GFSGS~\cite{jiang2025geometry} & & 0.40 & 0.59 & 0.39 & 0.38 & \sbest 0.72 & 0.59 & 0.65 & 1.08 & 0.93 & 0.59 & 0.50 & 0.67 & 0.34 & 0.47 & 0.40 & & 0.58 & 16.8m \\
 & PGSR~\cite{chen2024pgsr} && \sbest 0.34 & 0.54 & 0.44 & 0.37 & \tbest 0.78 & 0.57 & 0.49 & 1.06 & \sbest 0.63 & 0.59 & 0.47 & \sbest 0.50 & \best 0.30 & 0.37 & 0.34 &&  0.52 & 30.5m\\
& GeoSVR~\cite{li2025geosvr} && \best 0.32 & \tbest 0.51 & \tbest 0.30 & \tbest 0.33 & \best 0.71 & \tbest 0.48 & \best 0.42 & \best 1.03 & \best 0.62 & \tbest 0.56 & \tbest 0.33 & \best 0.46 & \best 0.30 & \tbest 0.34 & \best 0.32 & & \best 0.47 & 53.3m \\
& Ours (20k) && 0.38 & \best 0.50 & \best 0.27 & \best 0.31 & 0.80 & \best 0.43 & \best 0.42 & \sbest 1.04 & \tbest 0.64 & \best 0.52 & \best 0.31 & \tbest 0.56 & \best 0.30 & \best 0.31 & \sbest 0.33 && \best 0.47 & 15.0m\\
& Ours (30k) && \tbest 0.37 & \best 0.50 & \best 0.27 & \best 0.31 & 0.81 & \best 0.43 & \best 0.42 & \tbest 1.05 & \tbest 0.64 & \best 0.52 & \sbest 0.32 & 0.58 & \best 0.30 & \best 0.31 & \sbest 0.33 && \tbest 0.48 & 25.3m\\

 \hline
\end{tabular}
}
\label{tab:dtu_result}
\end{table*}

%% file: table/tnt.tex
\begin{table}[t]
\centering
\caption{\textbf{Quantitative comparison on the Tanks $\&$ Temples Dataset~\cite{Knapitsch2017}}. We report the F1-score and average optimization time. }
\vspace{-0.4cm}
\resizebox{0.98\linewidth}{!}{
\begin{tabular}{@{}ll|cccccc|cc@{}}
\toprule
\multicolumn{2}{c|}{} & Barn & Cat. & Cour. & Igna. & Meet. & Truc. & Mean & Time \\
\hline
\multirow{3}{*}{\rotatebox[origin=c]{90}{implicit}}
  & NeuS         & 0.29 & 0.29 & 0.17 & \sbest 0.83 & 0.24 & 0.45 & 0.38 & >24h \\
  & Geo-NeuS     & 0.33 & 0.26 & 0.12 & 0.72 & 0.20 & 0.45 & 0.35 & >24h \\
  & Neurlangelo  & \best 0.70 & 0.36 & \tbest 0.28 & \best 0.89 & 0.32 & 0.48 & 0.50 & >24h \\
\cline{2-2} \cline{3-10} 
\multirow{6}{*}{\rotatebox[origin=c]{90}{explicit}}
  & 2D GS        & 0.36 & 0.23 & 0.13 & 0.44 & 0.16 & 0.26 & 0.30 & \sbest{15.5m} \\
  & GOF          & 0.51 & 0.41 & \tbest 0.28 & 0.68 & 0.28 & 0.59 & 0.46 & 71.6m \\
  & RaDe-GS      & 0.49 & 0.36 & 0.27 & 0.72 & 0.27 & 0.61 & 0.45 & \best 12.1m \\
  & PGSR         & 0.66 & \tbest 0.44 & 0.20 & 0.81 & \tbest 0.33 & \sbest 0.66 & \tbest 0.52 & 42.9m \\
  & GeoSVR       & \tbest 0.68 & \sbest 0.49 & \sbest 0.34 & \sbest 0.83 & \sbest 0.37 & \sbest0.66 & \sbest 0.56 & 66.4m \\
  & Ours         & \best 0.70 & \best 0.56 & \best 0.38 & 0.81 & \best 0.42 & \best 0.70 & \best 0.60 & \tbest{32.1m} \\
\bottomrule
\end{tabular}
}
\label{tab:tnt}
\end{table}

%% file: table/ablation.tex
\begin{table}[t]
\centering
\caption{\textbf{Ablation study on Tanks \& Temples ~\cite{Knapitsch2017}.}
The normal consistency loss and the single-view geometric loss in PGSR have a similar formulation. We denote them as $L_n$. $L_{gc}$is the geometric consistency loss. `exposure' represents the exposure compensation module from PGSR. We toggle each term on/off (\checkmark/--) and report the resulting reconstruction accuracy.}
\vspace{-0.2cm}
\resizebox{0.9\columnwidth}{!}{
\begin{tabular}{c|cc|cccc}
  & PGSR & RaDe-GS & \multicolumn{4}{c}{Ours} \\ \hline
$L_{gc}$    & \checkmark & \checkmark & -- & \checkmark & \checkmark & \checkmark \\
$L_n$             & \checkmark & \checkmark & \checkmark & -- & \checkmark & \checkmark \\
exposure          & \checkmark & \checkmark & \checkmark & \checkmark & -- & \checkmark \\
F1-score          & 0.52 & 0.52 & 0.60 & 0.57 & 0.59 & 0.60 \\
\end{tabular}
}
\label{tab:ablation}
\end{table}

%% file: conclusion.tex
We reveal the intrinsic geometry for Gaussian Splatting. We regard Gaussian primitives as stochastic solids and design an appropriate attenuation function to make their volume rendering identical to their rasterization-based rendering. The stochastic theory enables depth rendering in a principled manner. Experiments show that our method outperforms state-of-the-art methods.

%% file: figure_only.tex
\begin{figure*}[t]
    \centering
    \includegraphics[width=\linewidth]{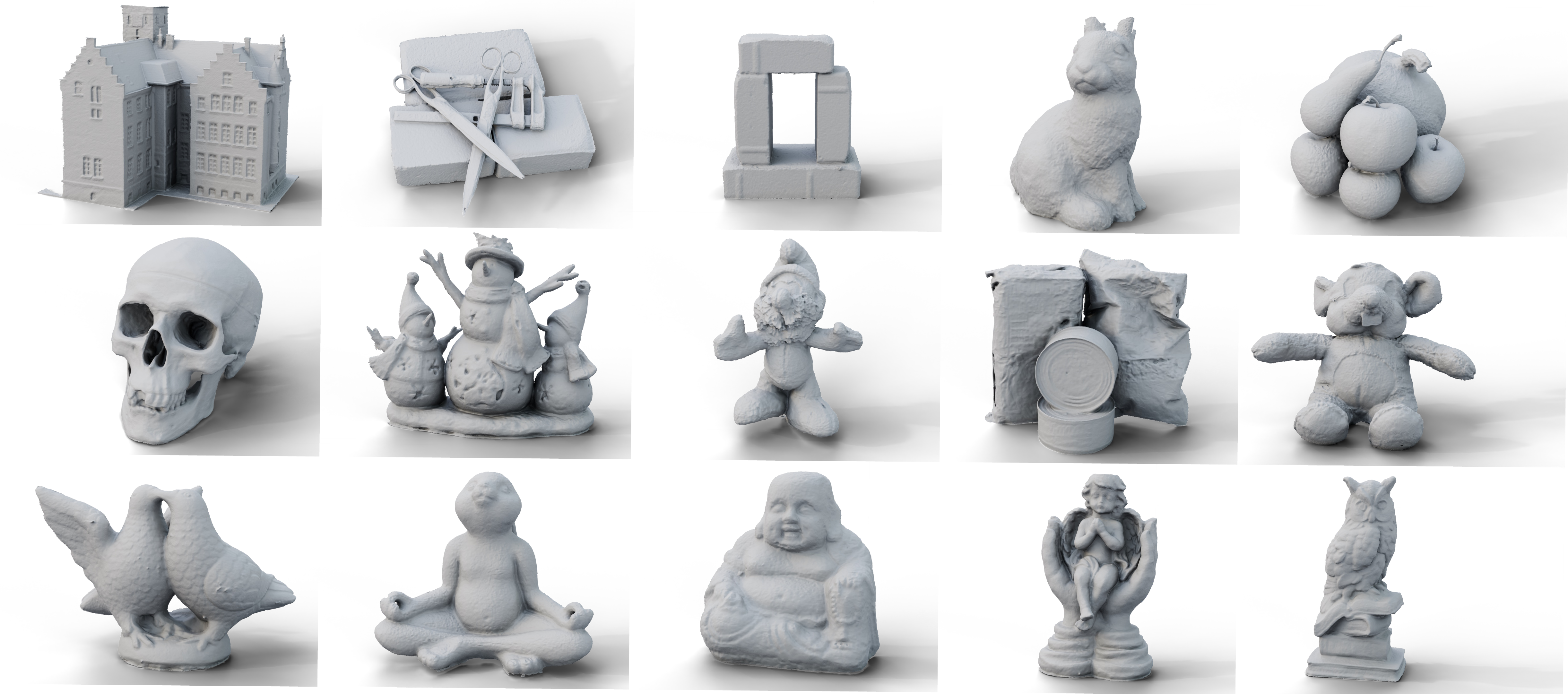}
    \caption{\textbf{Qualitative results on the DTU~\cite{jensen2014large} dataset.}}
    \label{fig:dtu}
\end{figure*}

\begin{figure*}[t]
    \centering
    \includegraphics[width=1.0\linewidth]{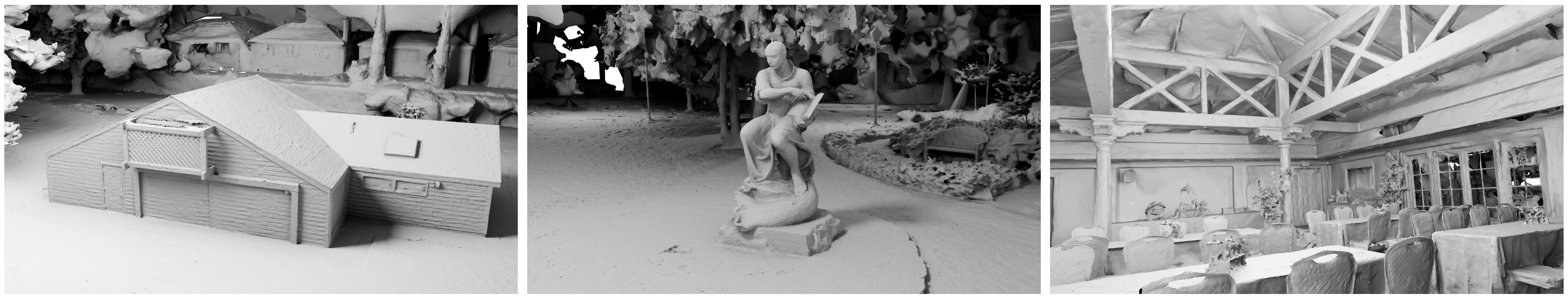}
    \caption{\textbf{Qualitative results on the Tanks \& Temples~\cite{Knapitsch2017} dataset.}}
    \label{fig:tnt}
\end{figure*}

\begin{figure*}
    \centering
    \includegraphics[width=1.0\linewidth]{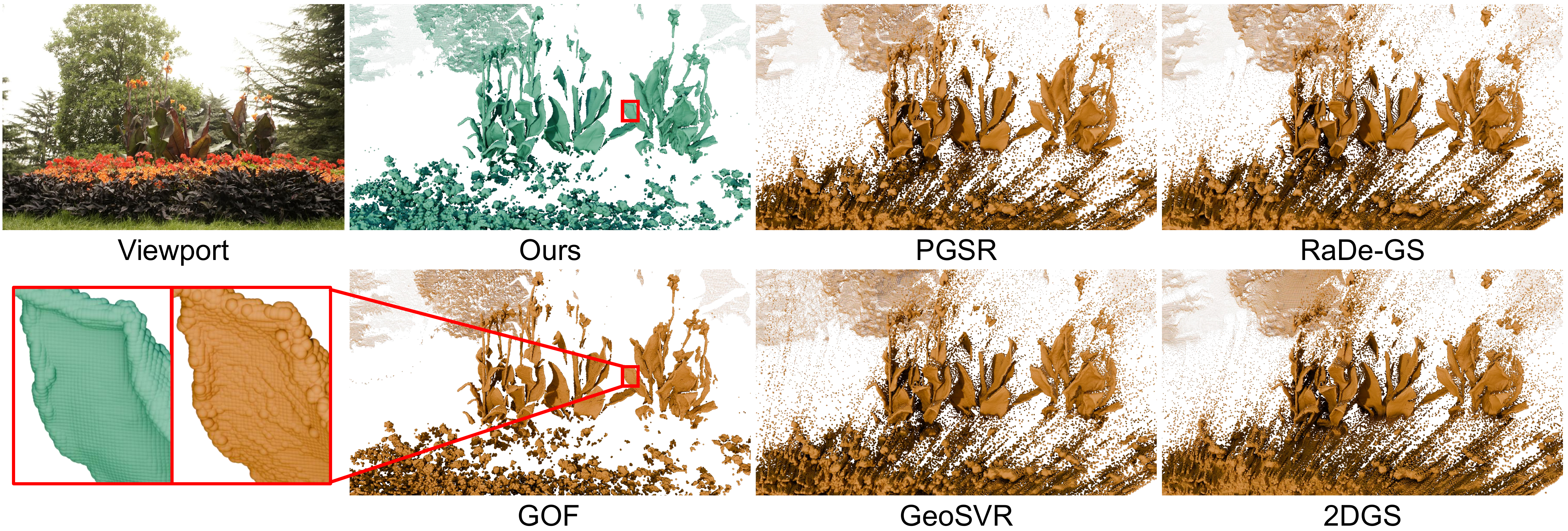}
    \caption{\textbf{Qualitative comparison of depth rendering among our method and prior methods.} 
    We visualize depth maps by back-projecting them into 3D point clouds.  RaDe-GS uses multi-view regularization and expected depth; 2DGS uses expected depth.}
    \label{fig:depth_figure_only}
\end{figure*}

\begin{figure*}
    \centering
    \setlength{\abovecaptionskip}{0.1cm}
    \includegraphics[width=1.0\linewidth]{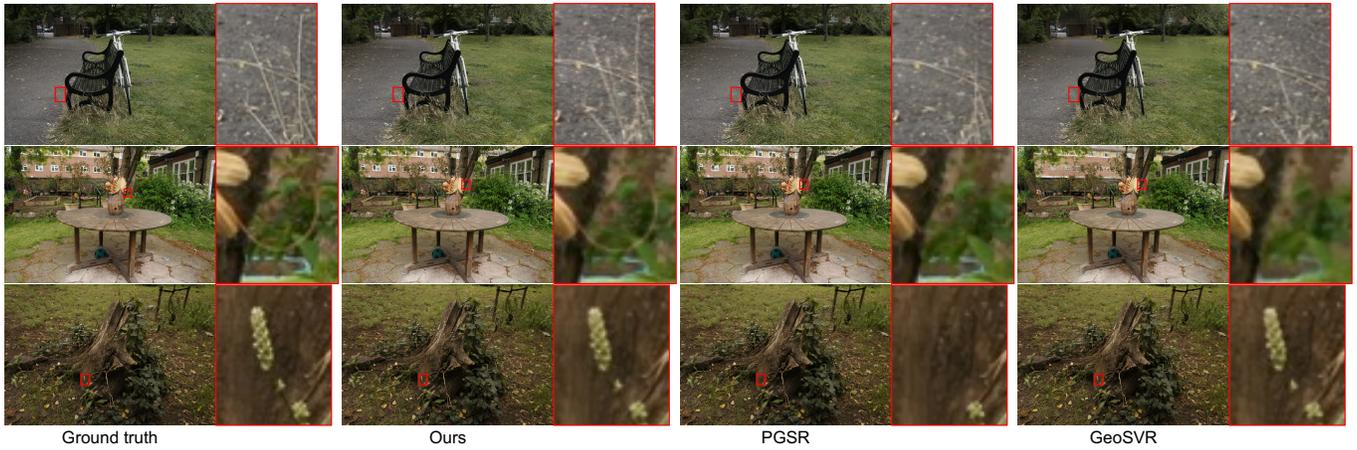}
    \caption{\textbf{Qualitative comparison of novel view synthesis among our method and prior methods on the Mip-NeRF360~\cite{barron2022mipnerf360} dataset.}}
    \label{fig:novel_view_synthesis}
\end{figure*}

\begin{figure*}[t]
    \centering
    \includegraphics[width=1.0\linewidth]{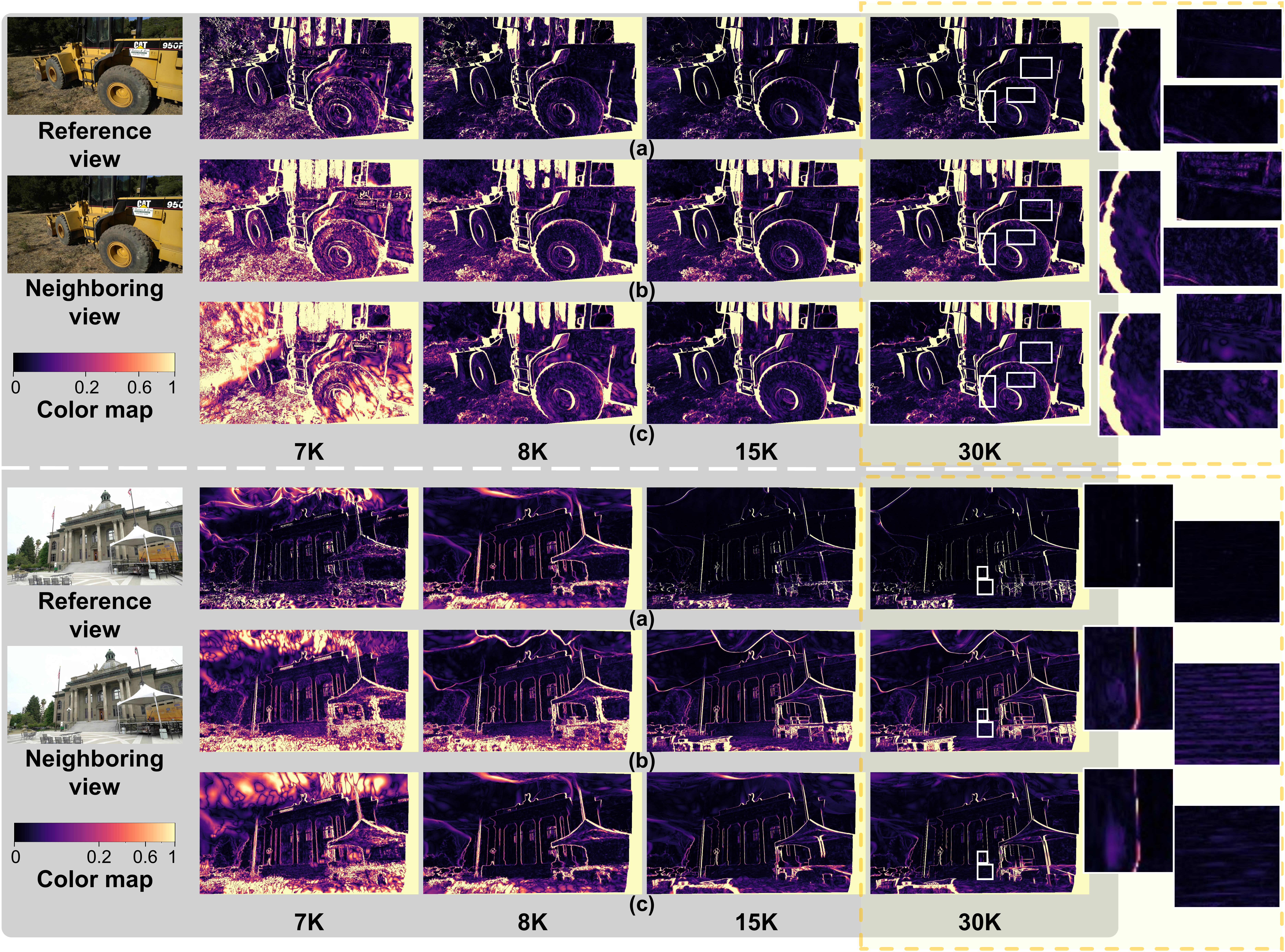}
    \caption{\textbf{Cycle reprojection error per iteration}. We visualize the cycle reprojection error between a reference view and its nearest neighboring view throughout optimization. We show the projection error of (a) our method, (b) PGSR, and (c) RaDe-GS with multi-view regularization. Zoomed-in patches are shown on the right.}
    \label{fig:reprojection_more}
\end{figure*}

%% file: supp.tex
\clearpage

\makeatletter
\newlength\ACMSuppTitlePreSkip
\newlength\ACMSuppTitlePostSkip
\setlength\ACMSuppTitlePreSkip{1.5\baselineskip}
\setlength\ACMSuppTitlePostSkip{2.5\baselineskip}

\newcommand{\ACMSupplementTitle}[1]{%
  \clearpage
  \begingroup
    \if@twocolumn
      \onecolumn
      \twocolumn[{%
        \vskip\ACMSuppTitlePreSkip
        \par\noindent
        \@titlefont #1\par
        \vskip\ACMSuppTitlePostSkip
      }]%
    \else
      \vskip\ACMSuppTitlePreSkip
      \par\noindent
      \@titlefont #1\par
      \vskip\ACMSuppTitlePostSkip
    \fi
  \endgroup
}
\makeatother

\ACMSupplementTitle{Supplementary: Geometry-Grounded Gaussian Splatting}

\section{Implementation of Depth Rendering}
In this section, we detail the implementation of the forward and backward passes for depth rendering.
\subsection{Forward Pass} 
We begin with the initial median depth $t_{init}$ obtained from RaDe-GS~\cite{zhang2024rade}. We then establish an initial depth interval $[t_{init} - r,\, t_{init} + r]$ and search for the median depth within this range, setting r to 0.4 during training. To perform the binary search, we need to traverse the Gaussians along the camera ray and record the transmittance at the mid-point, comparing it to the target value of 0.5. However, traversing Gaussians can be time-consuming, so we aim to reduce the number of Gaussian traversals. Specifically, instead of splitting the interval into two segments by a midpoint, we evenly divide it into eight segments using seven segment points and record the transmittance at each segment point. 
After each traversal, we locate the segment whose endpoint transmittance values fall on opposite sides of $0.5$ and use it as the new search interval.
Under this setting, a single traversal is equivalent to three binary-search iterations.
We repeat this process 5 times, gradually narrowing the interval until the final depth error is within $0.8\times8^{-5}=2.441\times10^{-5}$.
In the first pass, we also record the transmittance values at both ends of the interval, \ie, $t_{init} - r$, $t_{init} + r$. If both values are above 0.5 or below 0.5, we mask the pixel for geometric regularization.
\subsection{Backward Pass}
The backpropagation of depth can be calculated as,
\begin{equation}
    \frac{\partial L}{\partial t_{med}}\cdot\frac{\partial t_{med}}{\partial \theta},
\label{eq:chain}
\end{equation}
where $L$ denotes the loss, $t_{med}$ is the median depth computed in the forward pass, and $\theta$ represents the parameters of the Gaussians along the camera ray. We further write $\theta=\{\theta_i\}$, where $\theta_i$ denotes the parameters of the $i$-th Gaussian.
The first term $\partial L/\partial t_{med}$ is the input of the backward function, and we need to calculate the second term.
The formulation of the second term is shown in Equation~\ref{eq:depth_gradient} of the main paper and will be derived in section~\ref{sec:depth_gradient}. It is composed of $\partial T(t;\theta)/\partial t|_{t=t_{med}}$ and $\partial T(t_{med};\theta)/\partial \theta$. We traverse the Gaussians along the ray twice, computing the two terms in separate passes.

In the first pass, we calculate $\partial T(t;\theta)/\partial t|_{t=t_{med}}$.
Equation~\ref{eq:t_depth} and Equation~\ref{eq:T} of the main paper show that,
\begin{equation}
    T(t_{med};\theta) = \prod_i T_i(t_{med};\theta_i)=0.5.
\end{equation}
We can obtain:

\begin{equation}
\begin{aligned}
    \frac{\partial T(t;\theta)}{\partial t}|_{t=t_{med}} &= \sum_i \sum_{j\neq i} T_j(t_{med};\theta_j)\frac{\partial T_i(t;\theta_i)}{\partial t}|_{t=t_{med}}\\
    &= \sum_i \frac{T(t_{med};\theta)}{T_i(t_{med};\theta_i)}\frac{\partial T_i(t;\theta_i)}{\partial t}|_{t=t_{med}}\\
    &= \sum_i \frac{0.5}{T_i(t_{med};\theta_i)}\frac{\partial T_i(t;\theta_i)}{\partial t}|_{t=t_{med}}.
\label{eq:gradiant_T_t}
\end{aligned}
\end{equation}

After computing Equation~\ref{eq:gradiant_T_t}, we modify the standard Gaussian Splatting color-accumulation backward pass to additionally compute $\partial T(t_{med};\theta)/\partial \theta_i$ for each Gaussian.
\begin{equation}
    \begin{aligned}
    \frac{\partial T(t;\theta)}{\partial \theta_i} &= \sum_{j\neq i} T_j(t_{med};\theta_j)\frac{\partial T_i(t_{med};\theta_i)}{\partial \theta_i}\\
    &= \frac{T(t_{med};\theta)}{T_i(t_{med};\theta_i)}\frac{\partial T_i(t_{med};\theta_i)}{\partial \theta_i}\\
    &= \frac{0.5}{T_i(t_{med};\theta_i)}\frac{\partial T_i(t_{med};\theta_i)}{\partial \theta_i}.
\label{eq:gradiant_T_theta}
    \end{aligned}
\end{equation}

The $\partial T_i(t;\theta_i)/\partial t|_{t=t_{med}}$ in Equation~\ref{eq:gradiant_T_t} and $\partial T_i(t_{med};\theta_i)/\partial \theta_i$ in Equation~\ref{eq:gradiant_T_theta} can be easily derived from the closed formed formulation of $T_i$, \ie, Equation~\ref{eq:Ti} of the main paper.
Using Equation~\ref{eq:depth_gradient} of the main paper, we plug Equation~\ref{eq:gradiant_T_t} and Equation~\ref{eq:gradiant_T_theta} into Equation~\ref{eq:chain}. The gradients are backpropagated to each Gaussian as,
\begin{equation}
\frac{\partial L}{\partial t_{med}}\cdot\frac{\partial t_{med}}{\partial \theta_i}=\frac{\partial L}{\partial t_{med}}\cdot\frac{\partial T(t_{med};\theta)}{\partial \theta_i}/ (-\frac{\partial T(t;\theta)}{\partial t}|_{t=t_{med}}).
\end{equation}

\section{Gaussian Splatting Preliminaries}
Gaussian Splatting represents a scene as a set of 3D Gaussian primitives. A single primitive is parameterized by a center $\mathbf{x}_c\in\mathbb{R}^3$, an opacity $o\in[0,1]$, and a symmetric positive definite covariance $\mathbf{\Sigma}\in\mathbb{R}^{3\times 3}$,
\begin{equation}
G(\mathbf{x}) = o\,\exp\!\Big(-(\mathbf{x}-\mathbf{x}_c)^\top \mathbf{\Sigma}^{-1}(\mathbf{x}-\mathbf{x}_c)\Big).
\end{equation}

\paragraph{From 3D to screen-space Gaussians.}
To render efficiently, Gaussian Splatting rasterizes each 3D Gaussian as an elliptical 2D Gaussian on the image plane. Let the camera extrinsics map world coordinates to camera coordinates, and denote the world-to-camera rotation as $\mathbf{W}$. The covariance in the camera frame is
\begin{equation}
\mathbf{\Sigma}_{cam}=\mathbf{W}\mathbf{\Sigma}\mathbf{W}^\top.
\end{equation}
Let $\pi(\cdot)$ be the perspective projection and $\mathbf{u}_c=\pi(\mathbf{x}_{c,cam})$ be the projected center on the image plane. Using a local affine approximation of $\pi$ around $\mathbf{x}_{c,cam}$, the screen-space covariance is obtained via multiplying with a Jacobian matrix:
\begin{equation}
\mathbf{\Sigma}'_{2D}=\mathbf{J}\,\mathbf{\Sigma}_{cam}\,\mathbf{J}^\top
=\mathbf{J}\mathbf{W}\mathbf{\Sigma}\mathbf{W}^\top\mathbf{J}^\top,
\label{eq:ray_cov2D}
\end{equation}
where $\mathbf{J}\in\mathbb{R}^{2\times 3}$ is the Jacobian of the perspective projection evaluated at $\mathbf{x}_{c,cam}$.

\paragraph{Per-pixel alpha.}
Given a pixel location $\mathbf{u}\in\mathbb{R}^2$, the Gaussian contributes an opacity (alpha) determined by its screen-space ellipse:
\begin{equation}
\alpha(\mathbf{u}) = o\,e^{-(\mathbf{u}-\mathbf{u}_c)^\top \mathbf{\Sigma}_{2D}^{'-1}(\mathbf{u}-\mathbf{u}_c)},
\label{eq:alpha_supp}
\end{equation}
where $\mathbf{u}_c$ is the projected Gaussian center. In practice, each Gaussian is evaluated only on pixels within a finite screen-space support to keep rasterization fast.

\paragraph{Alpha compositing.}
For each pixel, let $N$ denote the set of Gaussians whose projected support overlaps that pixel. These Gaussians are depth-sorted and accumulated using standard alpha blending. Denoting the per-Gaussian color as $\mathbf{c}_i$ and $\alpha_i=\alpha_i(\mathbf{u})$, the rendered color is
\begin{equation}
\mathbf{C}(\mathbf{u})=\sum_{i\in N}\mathbf{c}_i\,\alpha_i \prod_{j=1}^{i-1}(1-\alpha_j).
\label{eq:alpha_blending}
\end{equation}

\section{Loss Functions}
During training, we employ three loss terms: the photometric loss from Gaussian Splatting~\cite{kerbl3Dgaussians}, the normal consistency loss from 2D GS~\cite{huang20242d}, and the multi-view regularization from PGSR~\cite{chen2024pgsr}. We describe each term below.

\paragraph{Photometric loss.}
We follow~\cite{kerbl3Dgaussians} and define the photometric loss as a weighted combination of an $L_1$ term and a D-SSIM term between the rendered image and the ground-truth image:
\begin{equation}
    L_c = (1-\lambda)\, L_1 + \lambda\, L_{SSIM},
\label{eq:Lc}
\end{equation}
where $\lambda$ is a hyperparameter.

\paragraph{Normal consistency.}
Photometric supervision alone is insufficient to constrain geometry, so we introduce additional geometric regularization. Specifically, we adopt the normal-consistency loss from 2D GS~\cite{huang20242d}, which encourages the Gaussian normals to agree with the surface normal estimated from the rendered depth map. Concretely, we compute a reference normal $\tilde{\mathbf{n}}$ by applying finite differences to the depth map and penalize its angular deviation from each Gaussian normal:
\begin{equation}
\mathcal{L}_{n} = \sum_{i} \omega_i \bigl(1-\mathbf{n}_i^\top \tilde{\mathbf{n}}\bigr),
\end{equation}
where $\mathbf{n}_i$ is the normal of the $i$-th Gaussian along the ray, and $\omega_i=\alpha_i\prod_{j=1}^{i-1}(1-\alpha_j)$ is its alpha-compositing weight.

\paragraph{Multi-view regularization.}
We adopt the multi-view regularization of PGSR~\cite{chen2024pgsr} to our method, which combines a photometric term with an explicit geometric cycle-consistency term. Concretely, for each reference pixel $\mathbf{u}_r$, we approximate the local surface as a plane using the rendered depth and normal, and use the induced plane homography to relate the reference view to a neighboring view:
\begin{equation}
\mathbf{H}_{rn} = \mathbf{K}_n \left( \mathbf{R}_{rn} + \frac{\mathbf{T}_{rn}\mathbf{n}_r^{\top}}{\mathbf{p}_r^{\top}\mathbf{n}_r} \right)\mathbf{K}_r^{-1},
\end{equation}
where $\mathbf{K}_r$ and $\mathbf{K}_n$ are the intrinsics, $(\mathbf{R}_{rn},\mathbf{T}_{rn})$ is the relative pose from the reference to the neighboring camera, $\mathbf{n}_r$ is the rendered normal at $\mathbf{u}_r$, and $\mathbf{p}_r$ is the 3D point in the reference camera frame obtained from the rendered depth along the ray through $\mathbf{u}_r$.

Using $\mathbf{H}_{rn}$, we warp the neighboring image into the reference frame and enforce patch-level photometric consistency via normalized cross-correlation (NCC):
\begin{equation}
L_{pc} = \sum_{\mathbf{u}_r} w(\mathbf{u}_r)\Big(1-\text{NCC}\big(I_r(\mathbf{u}_r),\, I_n(\mathbf{H}_{rn}\mathbf{u}_r)\big)\Big),
\end{equation}
where $I_r$ and $I_n$ denote the reference and neighboring images.
To handle occlusions and unreliable correspondences, PGSR defines a confidence weight from a forward--backward reprojection cycle.
Specifically, letting $\mathbf{H}_{nr}$ denote the homography that maps from the neighboring view back to the reference view, the cycle reprojection error is
\begin{equation}
\phi(\mathbf{u}_r)=\left\lVert \mathbf{u}_r - \mathbf{H}_{nr}\mathbf{H}_{rn}\mathbf{u}_r \right\rVert_2,
\end{equation}
which is the same reprojection error introduced in the main paper.
The confidence is then
\begin{equation}
w(\mathbf{u}_r)=
\begin{cases}
\exp\big(-\phi(\mathbf{u}_r)\big), & \phi(\mathbf{u}_r) < 1, \\
0, & \phi(\mathbf{u}_r) \ge 1,
\end{cases}
\end{equation}
thus discarding pixels with large cycle error.

In addition to the photometric term, PGSR directly penalizes the cycle reprojection error to encourage view-consistent geometry:
\begin{equation}
L_{gc} = \sum_{\mathbf{u}_r} w(\mathbf{u}_r)\,\phi(\mathbf{u}_r).
\end{equation}
The overall multi-view regularization is
\begin{equation}
L_{mv}=w_{pc}L_{pc}+w_{gc}L_{gc},
\end{equation}
where $w_{pc}$ and $w_{gc}$ control the relative strength of photometric and geometric consistency.

Our final training loss $\mathcal{L}$ is,
\begin{equation}
    \mathcal{L} = \mathcal{L}_{c} + w_n \mathcal{L}_{n} + L_{mv}.
\end{equation}
We use $w_n=0.05$, $w_{pc}=0.6$, $w_{gc}=0.02$, and set $\lambda=0.2$ in Equation~\ref{eq:Lc}.

\begin{figure*}
    \includegraphics[width=1.0\linewidth]{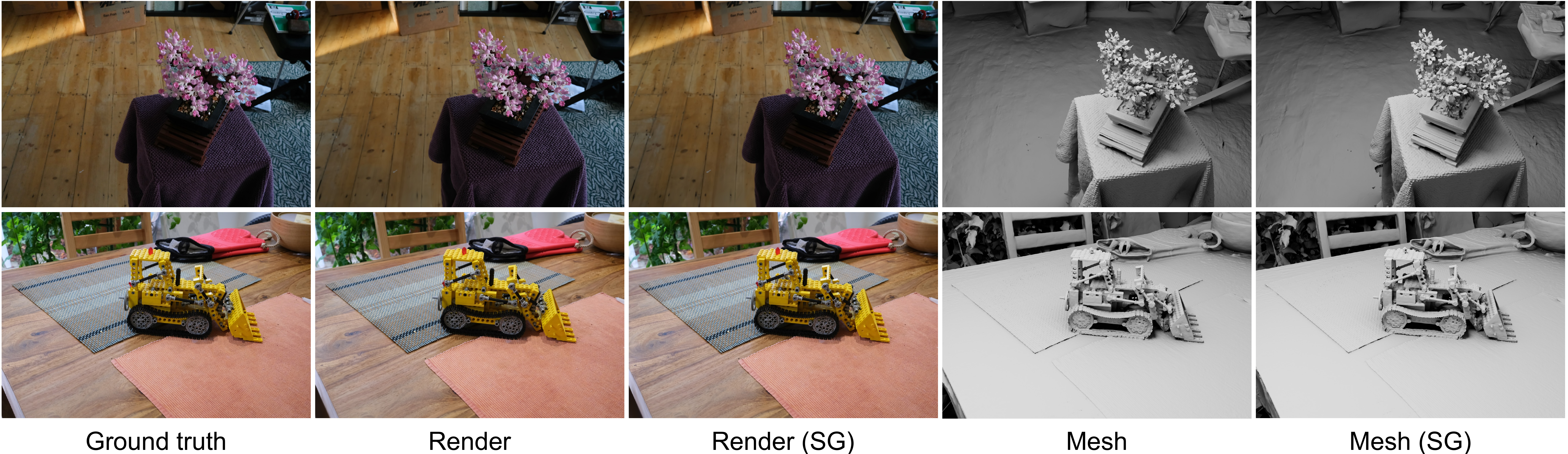}
    \caption{\textbf{Qualitative results on the Mip-NeRF 360 dataset~\cite{barron2022mipnerf360}.} We visualize novel view synthesis and shape reconstruction results for our method, and for our method augmented with the spherical Gaussian appearance model of RayGauss~\cite{blanc2025raygauss}. Incorporating spherical Gaussians improves rendering quality in specular regions.}
    \label{fig:mip360}
\end{figure*}
\section{Comparison on Novel View Synthesis}
We further compare novel view synthesis results across methods. Table~\ref{tab:360} shows the quantitative results on the Mip-NeRF 360 dataset. Our method achieves competitive performance compared with existing surface reconstruction baselines, while RayGaussX produces the overall best rendering quality. To isolate the effect of specular modeling, we augment our model with the spherical Gaussian mixture used in RayGaussX, denoted as Ours (SG). Notably, we use these Gaussian lobes only in this experiment; all other experiments use our default model without Gaussian lobes.
Figure~\ref{fig:mip360} shows the qualitative results of our method.

\input{table/mip360}

\section{Limitation and Future Work}
To maintain the efficiency of Gaussian Splatting, this paper only considers the volumetric effects when rendering depth. Future works can combine the stochastic theory with existing volume rendering methods~\cite{blanc2025raygaussx,kv2025stochasticsplats} to fully utilize the volumetric nature of stochastic when rendering color and normal maps. We believe this will lead to further improvement in shape reconstruction.

To compute the median depth, our binary search is initialized with a fixed depth interval. This interval must be sufficiently wide, which increases the number of search steps and slows training. For large-scale scenes, the true median depth may even fall outside this preset range, hindering effective optimization. We leave it to future work to develop adaptive bracketing strategies that reliably locate the median and tighten the initial interval, further accelerating depth rendering.

In Marching Tetrahedra, while the vertex placement is Gaussian-aware, the subsequent 3D Delaunay triangulation step remains general-purpose. In practice, reconstructing thin or near-planar structures often requires dense vertices. Designing Gaussian-specific tetrahedralization and refinement strategies is a meaningful direction for future work.

As we have bridged the gap between Gaussian and NeRF reconstruction methods, future work could consider adopting geometric regularization from NeRF-based methods, \eg, Neuralangelo~\cite{li2023neuralangelo}, for Gaussian Splatting-based methods to enhance shape quality.

\section{Proof of the Equation~\ref{eq:G2v} in the main paper}
\begin{figure}[h]
    \centering
    \includegraphics[width=1.0\linewidth]{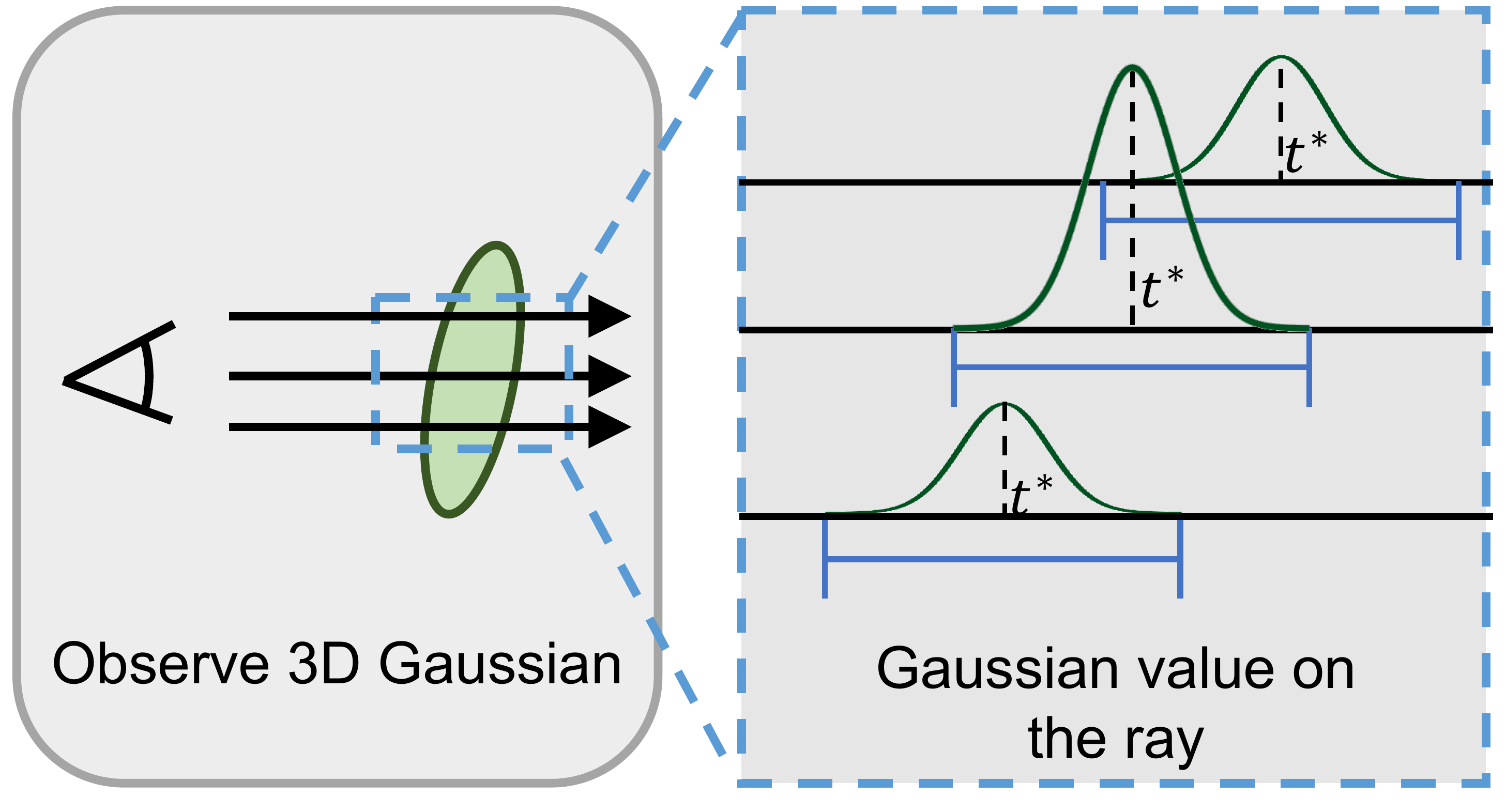}
    \caption{Gaussian Splatting uses local affine approximation so that rays are parallel to each other. Gaussian functions on the rays have the same variance but different maximum values.}
    \label{fig:maximum_supp}
\end{figure}

As shown in Figure~\ref{fig:maximum_supp}, Gaussian Splatting~\cite{kerbl3Dgaussians} applies a local affine approximation when projecting a Gaussian primitive. As a result, the light rays from the camera center are parallel to each other. The 3D Gaussian values on each ray form a 1D Gaussian function, which is denoted as $G_{uv}(t)$.
It is interesting to notice that these 1D Gaussians on different light rays have the same variance but different maximum values. 

Gaussian Splatting performs volume rendering on a single primitive as:
\begin{equation}
        G'_{2D}(u,v) = \int_{-\infty}^{+\infty} G_{uv}(t)\, dt,
    \label{eq:ray_integration}
\end{equation}
which is proportional to the maximum value of $G_{uv}(t)$ because of the identical spread, shown in Figure~\ref{fig:maximum_supp}.

Furthermore, as shown in Equation~\ref{eq:alpha_supp}, Gaussian Splatting normalizes the maximum value of the 2D Gaussian to opacity $o$, which aligns the maximum value between the 2D and the 3D Gaussians, ensuring that \textbf{the 2D Gaussian value matches the maximum value of the corresponding 1D Gaussian along the ray}.
This conclusion facilitates the derivation of the stochastic Gaussian solid.
While our result is derived from local affine projection, our method can be easily extended to perspective projection.

\section{Proof of "Gaussians as Stochastic Solids"}
Miller \etal~\shortcite{Miller:VOS:2024} propose a method to render stochastic opaque solids by converting the vacancy to the attenuation coefficient:
\begin{equation}
     \sigma(\mathbf{x}, \mathbf{\omega}) = \lvert \mathbf{\omega} \cdot \nabla log(\mathrm{v}(\mathbf{x})) \rvert=\frac{\lvert \mathbf{\omega} \cdot \nabla \mathrm{v}(\mathbf{x}) \rvert}{\mathrm{v}(\mathbf{x})},
     \label{eq:oav_supp}
 \end{equation}
 where $\mathbf{v}$ represents the vacancy of the stochastic solid.
 
In this section, we will prove that given a Gaussian primitive $G(\mathbf{x})$ rendered by Gaussian Splatting, we can find a solid that generates the same rendering results by the stochastic theory.
The vacancy of the solid should satisfy:
\begin{equation}
    \mathrm{v}(\mathbf{x}) = \sqrt{1-G(\mathbf{x})}.
    \label{eq:G2v_supp}
\end{equation}

We will prove that under the following constraints, the stochastic solid can be uniquely determined:
\begin{itemize}
    \item[--] When $G(\mathbf{x_1})\geq G(\mathbf{x_2})$, it follows that  $\mathrm{o}(\mathbf{x}_1)\geq\mathrm{o}(\mathbf{x}_2)$, indicating that positions closer to the Gaussian center have higher occupancy;
    \item[--] The occupancy of the solid approaches 0 when $\mathbf{x}$ is far from Gaussian center;
    \item[--] The occupancy $\mathrm{o}(\mathbf{x})$ is differentiable with respect to $\mathbf{x}$.
\end{itemize}

\textit{Proof}:
Assume a line $l$ parameterized by $t$ passes through $G(\mathbf{x})$. We get the value of $G(\mathbf{x})$ on the line forming a 1D Gaussian function $G(t)$, where
$t$ goes from $-\infty$ to $+\infty$ and reaches the maximum of the 1D Gaussian at $t^*$. 

Firstly, we derive the color from volume rendering.
According to the first assumption, the vacancy function along this line $l$ has the opposite monotonicity compared to the Gaussian function.
We will get the attenuation coefficient from Equation~\ref{eq:oav_supp}:
\begin{equation}
    \begin{aligned}
        \sigma(t) &=  \lvert \mathbf{\omega} \cdot \nabla log(\mathrm{v}(\mathbf{x})) \rvert\\
        &= \lvert \frac{\partial log(\mathrm{v}(\mathbf{x}))}{\partial t} \rvert\\
        &= \begin{cases}
        -\frac{\partial log(\mathrm{v}(\mathbf{x}))}{\partial t},\quad &t\leq t^* \\
        \frac{\partial log(\mathrm{v}(\mathbf{x}))}{\partial t},\quad &t>t^*
        \end{cases}
    \end{aligned}
    \label{eq:sigma}
\end{equation}

Since a Gaussian kernel has a uniform color, we can simplify the volume rendering:
 \begin{equation}
 \begin{aligned}
     \mathbf{C} &= \int_{t=-\infty}^{t=+\infty} T(t)\sigma(\mathbf{x}(t),\mathbf{\omega})\mathbf{c} \, dt\\
     &= \mathbf{c}\int_{t=-\infty}^{t=+\infty} T(t)\sigma(\mathbf{x}(t),\mathbf{\omega}) \, dt\\
     &=\mathbf{c}\int_{t=-\infty}^{t=+\infty} \, -d T(t)\\
     & = \mathbf{c} T(t)\big|_{t={+\infty}}^{t={-\infty}}=\mathbf{c}(1 - T(+\infty))
 \end{aligned}
 \label{eq:signle_color_rendering}
 \end{equation}
 We then substitute the Equation~\ref{eq:sigma} into Equation~\ref{eq:signle_color_rendering} to get the form of color from volume rendering:
 \begin{equation}
     \begin{aligned}
         T(\infty) &= T(-\infty, t^*) \times T(t^*, +\infty)\\
         &=e^{-\int_{-\infty}^{t^*} \sigma(\mathbf{x}(s),\mathbf{\omega})} \times e^{-\int_{t^*}^{+\infty} \sigma(\mathbf{x}(s),\mathbf{\omega})}\\
         & = e^{-(-log(\mathrm{v}(t))\big|_{-\infty}^{t^*})} \times e^{-(log(\mathrm{v}(t))\big|_{t^*}^{+\infty})}\\
         & = \frac{\mathrm{v}(t^*)}{\mathrm{v}(-\infty)} \times  \frac{\mathrm{v}(t^*)}{\mathrm{v}(+\infty)}\\
         & = \mathrm{v}(t^*)^2\\
        \mathbf{C} = &\mathbf{c}(1 - T(+\infty)) = \mathbf{c}(1 - \mathrm{v}(t^*)^2),
     \end{aligned}
     \label{eq:T_inf}
 \end{equation}
 where we use the second assumption that $\mathrm{v}(\infty)=1-\mathrm{o}(\infty) = 1$.
 
 Secondly, with the color derived from Gaussian Splatting and volume rendering, we can find the relationship between $\mathrm{v}(t^*)$ and $G(t^*)$:
 \begin{equation}
     \begin{aligned}
        \mathbf{c}G(t^*) =\mathbf{c}(1 - \mathrm{v}(t^*)^2)  
         \Rightarrow \mathrm{v}(t^*) = \sqrt{1 - G(t^*)}
     \end{aligned}
     \label{eq:maximum_relationship}
 \end{equation}

 Finally, we will generalize Equation~\ref{eq:maximum_relationship} from maximum points to all 3D points.  Different lines $l$ have different maximum points, and Equation~\ref{eq:maximum_relationship} should hold for the maximum point on any line.
 Given any $\mathbf{x}\in\mathbf{R}^3$, we can always find the direction $\mathbf{\omega}\in \mathbf{S}^2$ satisfying $\mathbf{\omega}\cdot\nabla G(\mathbf{x})=\frac{\partial G(\mathbf{x})}{\partial \omega}=0$, indicating that $\mathbf{x}$ is the maximum point along ray $l:\mathbf{x}+t\mathbf{\omega}$.
 Therefore, the equation should hold for any position $\mathbf{x}$, which reaches the unique solution of vacancy in Equation~\ref{eq:G2v_supp}.

 \section{Derivation of Equation~\ref{eq:Ti} of the main paper}
 In this section, we will derive the closed form $T_i(t)$ in Equation~\ref{eq:Ti} of the main paper, which is also the negative integral of the free-flight distribution $-\int p(t)\, dt$. For brief notation, we use $T$ to denote the transmittance of a single Gaussian.

 We start from $t_n = -\infty$.
 Similar to Equation~\ref{eq:T_inf}, when $t \textless t^*$,
 \begin{equation}
     T(-\infty, t) =e^{-\int_{-\infty}^{t^*} \sigma(\mathbf{x}(s),\mathbf{\omega})} = \mathrm{v}(t).
 \end{equation}
 When $t \textgreater t^*$,
  \begin{equation}
  \begin{aligned}
     T(-\infty, t) &= T(-\infty, t^*) \times e^{-\int_{-\infty}^{t^*} \sigma(\mathbf{x}(s),\mathbf{\omega})}\\
     &= \mathrm{v}(t^*) \times \frac{\mathrm{v}(t^*)}{\mathrm{v}(t)}\\
     &= \frac{\mathrm{v}(t^*)^2}{\mathrm{v}(t)}.
  \end{aligned}
 \end{equation}
In most cases, the Gaussian primitive is far from the camera, so we can simply use $T(-\infty,t)$ as $T(t)$.

\section{Derivation of Equation~\ref{eq:depth_gradient} of the main paper}\label{sec:depth_gradient}
In this section, we will derive the gradient of the depth $t_{med}$ with respect to the parameters of all the Gaussians along the ray.
Since $T(t_{med})$ is a constant value of 0.5, its differential is 0.
\begin{align}
    T(t_{med};\theta) &\equiv 0.5,\\
    dT(t_{med};\theta) &\equiv 0,
\end{align}
where $\theta$ represents the parameters of Gaussians along the ray.
We then expand the $dT$ and plug $t_{med}$ into it to derive the gradient:
\begin{align}
    dT(t;\theta) &= \frac{\partial T}{\partial t}dt+\frac{\partial T}{\partial \theta} \cdot d\theta\\
    0 &= \frac{\partial T}{\partial t}dt_{med}+\frac{\partial T}{\partial \theta} \cdot d\theta\\
    dt_{med} &= (-\frac{\partial T}{\partial \theta}/\frac{\partial T}{\partial t})\cdot d\theta.
\end{align}
So that the gradient of depth is derived as,
\begin{equation}
    \frac{\partial t_{med}}{\partial \theta} = -\frac{\partial T(t_{med};\theta)}{\partial \theta}/\frac{\partial T(t;\theta)}{\partial t}\big|_{t=t_{med}},
\end{equation}

%% file: table/mip360.tex
\begin{table}
\caption{\textbf{Quantitative results on Mip-NeRF 360 dataset.} The best scores are highlighted with colors.}
\label{tab:360}
\resizebox{.98\columnwidth}{!}{
\setlength\tabcolsep{3pt}
\begin{tabular}{@{}ll|ccc|ccc}
\toprule
                                                         &                        & \multicolumn{3}{c|}{Outdoor Scene}                     & \multicolumn{3}{c}{Indoor Scene}                       \\
                                                         &                        & PSNR $\uparrow$ & SSIM $\uparrow$ & LPIPS $\downarrow$ & PSNR $\uparrow$ & SSIM $\uparrow$ & LPIPS $\downarrow$ \\ \midrule
\multirow{6}{*}{\rotatebox[origin=c]{90}{\small NVS}}           & NeRF                   & 21.46           & 0.458           & 0.515              & 26.84           & 0.790           & 0.370              \\
                                                         & Deep Blending          & 21.54           & 0.524           & 0.364              & 26.40           & 0.844           & 0.261              \\
                                                         & Instant NGP            & 22.90           & 0.566           & 0.371              & 29.15           & 0.880           & 0.216              \\
                                                         & Mip-NeRF 360           & 24.47           & 0.691           & 0.283              & \tbest 31.72           & 0.917           & 0.180              \\
                                                         & 3DGS                   & 24.67           & 0.728           & 0.240              & 30.96           & 0.924           & 0.187              \\
                                                         & SVRaster               & 24.68           & 0.738           & 0.206              & 30.65           & 0.927           & 0.161              \\
                                                         & RayGaussX               & \best 25.24           & \best 0.761           & \best 0.167              & \best 32.43           & \best 0.943           & \best 0.146              \\\midrule
\multirow{7}{*}{\rotatebox[origin=c]{90}{\small Surface Recon.}} & BakedSDF               & 22.47           & 0.585           & 0.349              & 27.06           & 0.836           & 0.258              \\
                                                         & SuGaR                  & 22.93           & 0.629           & 0.356              & 29.43           & 0.906           & 0.225              \\
                                                         & 2DGS                   & 24.34           & 0.717           & 0.246              & 30.40           & 0.916           & 0.195              \\
                                                         & GOF                    &  24.82     & 0.750    & 0.202        & 30.79     & 0.924    & 0.184       \\
                                                         & VCR-GauS               & 24.31           & 0.707           & 0.280              &  30.53    & 0.921    & 0.184       \\
                                                         & PGSR                   &  24.76    & 0.752     & 0.203       & 30.36           & \tbest 0.934     & \sbest 0.147        \\ 
                                                         & GeoSVR  & 24.83    & 0.738    & 0.218       & 30.46    & 0.921    & 0.172\\
                                                           & Ours (SG)  &  \tbest 24.97   &  \tbest 0.754    &  \tbest 0.200       & \sbest 32.18    & \sbest 0.938    &  \tbest 0.150\\
                                                            & Ours   &  \sbest 25.09  &  \sbest 0.760    & \sbest 0.196       & 31.02    &  \tbest 0.934   &  0.154\\\bottomrule
\end{tabular}%
}
\end{table}